\documentclass{article}

\PassOptionsToPackage{numbers}{natbib}

\usepackage[final]{neurips_2024}

\usepackage[utf8]{inputenc} % allow utf-8 input
\usepackage[T1]{fontenc}    % use 8-bit T1 fonts
\usepackage{hyperref}       % hyperlinks
\usepackage{url}            % simple URL typesetting
\usepackage{booktabs}       % professional-quality tables
\usepackage{amsfonts}       % blackboard math symbols
\usepackage{nicefrac}       % compact symbols for 1/2, etc.
\usepackage{microtype}      % microtypography
\usepackage[table,xcdraw]{xcolor}

\usepackage{inconsolata}

\usepackage{amsmath}
\usepackage{bbm}
\usepackage{amsfonts}
\usepackage{amssymb}
\usepackage{graphicx}
\usepackage{enumitem}
\usepackage{booktabs} % For formal tables
\usepackage{xcolor}
\usepackage{listings}
\usepackage{chngpage}
\usepackage{tcolorbox}
\usepackage{lipsum}
\usepackage{array} 
\usepackage{ragged2e}
\usepackage{lipsum}
\usepackage{multirow}
\usepackage{wrapfig}
\usepackage{float}

\newcommand{\aref}[1]{\hyperref[#1]{Appendix~\ref*{#1}}}

\definecolor{mygreen}{HTML}{008000}
\definecolor{olive}{HTML}{baba65}
\definecolor{beige}{HTML}{d4b996}

\definecolor{syntaxerror}{rgb}{0.890, 0.102, 0.110}
\definecolor{wrongentrelpropertytype}{rgb}{0.651, 0.808, 0.890}
\definecolor{reverseddirection}{rgb}{0.122, 0.471, 0.706}
\definecolor{entitylinking}{rgb}{0.698, 0.875, 0.541}
\definecolor{schemaviolation}{rgb}{0.984, 0.604, 0.600}
\definecolor{patternnotalignedwithquestion}{rgb}{0.992, 0.749, 0.435}
\definecolor{incorrectgrouping}{rgb}{0.792, 0.698, 0.839}
\definecolor{incorrectfiltering}{rgb}{0.416, 0.239, 0.604}
\definecolor{resultsnotalignedwithquestion}{rgb}{0.9, 0.9, 0.500}
\definecolor{incorrectdeduplication}{rgb}{0.694, 0.349, 0.157}

\newtcbox{\whitebg}[1][]{%
    on line, % Ensures it stays on the same line
    colback=white, % Background color
    colframe=white, % Frame color (same as background)
    arc=1mm, % Rounded corners
    boxrule=0pt, % No border
    left=0pt, right=1pt, top=8pt, bottom=4pt, % Adjust padding
    height=2\baselineskip, % Set height to one row (baselineskip)
    valign=center, % Vertically center the text
    #1 % Allow additional customization
}

\newtcbox{\rprop}[1][]{%
    on line, % Ensures it stays on the same line
    fontupper=\color{darkgray}\ttfamily,
    colback=olive, % Background color
    colframe=olive, % Frame color (same as background)
    arc=1mm, % Rounded corners
    boxrule=1pt, % No border
    left=1pt, right=1pt, top=4pt, bottom=4pt, % Adjust padding
    height=1.2\baselineskip, % Set height to one row (baselineskip)
    valign=center, % Vertically center the text
    #1 % Allow additional customization
}

\newtcbox{\nprop}[1][]{%
    on line, % Ensures it stays on the same line
    fontupper=\color{darkgray}\ttfamily,
    colback=beige, % Background color
    colframe=beige, % Frame color (same as background)
    arc=1mm, % Rounded corners
    boxrule=1pt, % No border
    left=1pt, right=1pt, top=4pt, bottom=4pt, % Adjust padding
    height=1.2\baselineskip, % Set height to one row (baselineskip)
    valign=center, % Vertically center the text
    #1 % Allow additional customization
}

\newtcbox{\rlabel}[1][]{%
    on line, % Ensures it stays on the same line
    fontupper=\color{olive}\ttfamily,
    colback=white, % Background color
    colframe=olive, % Frame color (same as background)
    arc=1mm, % Rounded corners
    boxrule=0.5pt, % No border
    left=1pt, right=1pt, top=4pt, bottom=4pt, % Adjust padding
    height=1.2\baselineskip, % Set height to one row (baselineskip)
    valign=center, % Vertically center the text
    #1 % Allow additional customization
}

\newtcbox{\nlabel}[1][]{%
    on line, % Ensures it stays on the same line
    fontupper=\color{beige}\ttfamily,
    colback=white, % Background color
    colframe=beige, % Frame color (same as background)
    arc=1mm, % Rounded corners
    boxrule=0.5pt, % No border
    left=1pt, right=1pt, top=4pt, bottom=4pt, % Adjust padding
    height=1.2\baselineskip, % Set height to one row (baselineskip)
    valign=center, % Vertically center the text
    #1 % Allow additional customization
}

\lstdefinestyle{TinyCypher}{
    language=SQL, % Use SQL as a base language since Cypher is similar
    morekeywords={
        MATCH, RETURN, WHERE, CREATE, DELETE, DETACH, SET, MERGE, ON, 
        OPTIONAL, WITH, DISTINCT, AS, LIMIT, ORDER, BY, SKIP, ASC, DESC, 
        AND, OR, NOT, IN, IS, STARTS, ENDS, CONTAINS, TRUE, FALSE, NULL, CALL, UNWIND
    }, % Add Cypher-specific keywords
    basicstyle=\ttfamily\tiny, % Code font style
    keywordstyle=\color{mygreen}\bfseries, % Keywords in bold blue
    stringstyle=\color{purple}, % String literals in orange
    commentstyle=\color{gray}\itshape, % Comments in italic gray
    numbers=none, % Line numbers on the left
    numberstyle=\tiny\color{gray}, % Line numbers style
    stepnumber=1, % Show line numbers for each line
    numbersep=10pt, % Distance from code to line numbers
    tabsize=2, % Tab size
    captionpos=b, % Position of the caption (b for bottom)
    breaklines=true, % Enable line breaking
    breakatwhitespace=true, % Allow breaking at whitespaces
    showspaces=false, % Don't show spaces
    showstringspaces=false, % Don't show spaces in strings
    escapeinside={(*@}{@*)}, % For escaping to LaTeX within the code
    frame=none, % Remove the frame around the code
    aboveskip=0pt, % Reduce space above the listing
    belowskip=0pt, % Reduce space below the listing
    extendedchars=true,
    literate={á}{{\'a}}1 {ã}{{\~a}}1 {é}{{\'e}}1 {í}{{\'i}}1 {ó}{{\'o}}1 {ú}{{\'u}}1 {ç}{{\c{c}}}1,
}

\lstdefinestyle{TinyJSON}{
    language=,
    basicstyle=\ttfamily\tiny, % Code font style
    keywordstyle=\color{mygreen}\bfseries, % Keywords in bold blue
    stringstyle=\color{purple}, % String literals in orange
    commentstyle=\color{gray}\itshape, % Comments in italic gray
    numbers=none, % Line numbers on the left
    numberstyle=\tiny\color{gray}, % Line numbers style
    stepnumber=1, % Show line numbers for each line
    numbersep=10pt, % Distance from code to line numbers
    tabsize=2, % Tab size
    captionpos=b, % Position of the caption (b for bottom)
    breaklines=true, % Enable line breaking
    breakatwhitespace=true, % Allow breaking at whitespaces
    showspaces=false, % Don't show spaces
    showstringspaces=false, % Don't show spaces in strings
    escapeinside={(*@}{@*)}, % For escaping to LaTeX within the code
    frame=none, % Remove the frame around the code
    aboveskip=0pt, % Reduce space above the listing
    belowskip=0pt, % Reduce space below the listing
    extendedchars=true,
    literate={á}{{\'a}}1 {ã}{{\~a}}1 {é}{{\'e}}1 {í}{{\'i}}1 {ó}{{\'o}}1 {ú}{{\'u}}1 {ç}{{\c{c}}}1,
}

\newcommand{\stitle}[1]{\vspace{0.2em}\noindent\textbf{#1}}%%
\newcommand{\hide}[1]{}%%
\newcommand{\eg}{{\itshape e.g.}, }%%
\newcommand{\ie}{{\itshape i.e.}, }%%

\newcommand{\sajhidden}[1]{\textcolor{red}{}} %Sajjadur Comment: #1

\title{{\scshape CypherBench}: Towards Precise Retrieval over Full-scale Modern Knowledge Graphs in the LLM Era}

\author{%
 Yanlin Feng$^\alpha$ \quad Simone Papicchio\thanks{The work began during Simone Papicchio's internship at Megagon Labs. As part of one subtask of his overall internship goal, he implemented an initial version of the benchmark that involved SQL-inspired template design, query categorization, and validation of the generated benchmark. The work has since further evolved to broaden and bolster the template generation process and redefining query categories while introducing new evaluation metrics.}~~$^{\beta \gamma}$ \quad Sajjadur Rahman$^\alpha$ \vspace{0.3em} \\ 
  $^\alpha$Megagon Labs \ \ $^\beta$Politecnico di Torino \ \ $^\gamma$EURECOM\\
  \texttt{\{yanlin, sajjadur\}@megagon.ai} \ \ \texttt{simone.papicchio@polito.it} 
}
\begin{document}

\maketitle

% \begin{abstract}
%   Retrieval from graph data is crucial for augmenting large language models (LLM) with both open-domain knowledge and private enterprise data, and it is also a key component in the recent GraphRAG system \cite{edge2024local}. Despite decades of research on knowledge graphs and knowledge base question answering, leading LLM frameworks (\eg Langchain and LlamaIndex) have only minimal support for retrieval from modern encyclopedic knowledge graphs like Wikidata. In this paper, we analyze the root cause and suggest that modern RDF knowledge graphs (\eg Wikidata, Freebase) are less efficient for LLMs due to overly large schemas that far exceed the typical LLM context window, use of resource identifiers, overlapping and ambiguous relation types and lack of normalization. As a solution, we propose \textit{property graph views} on top of the underlying RDF graph that can be efficiently queried by LLMs using \textit{Cypher}.  We instantiated this idea on Wikidata and introduced {CypherBench}, the first benchmark with 11 large-scale, multi-domain property graphs with 7.8 million entities and over 10,000 questions. To achieve this, we tackled several key challenges, including developing an RDF-to-property graph conversion engine, creating a systematic pipeline for text-to-Cypher task generation, and designing new evaluation metrics. 
%   % We also open-source a comprehensive set of tools to facilitate future research in large-scale graph retrieval. \yanlin{Github and Huggingface link}
% \end{abstract}

\vspace{-0.8em}

\begin{abstract}
    Retrieval from graph data is crucial for augmenting large language models (LLM) with both open-domain knowledge and private enterprise data, and it is also a key component in the recent GraphRAG system \cite{edge2024local}. Despite decades of research on knowledge graphs and knowledge base question answering, leading LLM frameworks (\eg Langchain and LlamaIndex) have only minimal support for retrieval from modern encyclopedic knowledge graphs like Wikidata. In this paper, we analyze the root cause and suggest that modern RDF knowledge graphs (\eg Wikidata, Freebase) are less efficient for LLMs due to overly large schemas that far exceed the typical LLM context window, use of resource identifiers, overlapping relation types and lack of normalization. As a solution, we propose \textit{property graph views} on top of the underlying RDF graph that can be efficiently queried by LLMs using \textit{Cypher}.  We instantiated this idea on Wikidata and introduced {CypherBench}, the first benchmark with 11 large-scale, multi-domain property graphs with 7.8 million entities and over 10,000 questions. To achieve this, we tackled several key challenges, including developing an RDF-to-property graph conversion engine, creating a systematic pipeline for text-to-Cypher task generation, and designing new evaluation metrics. 
\begin{center}

\begin{tabular}{ll}
    \raisebox{-1.5pt}{\includegraphics[height=1.05em]{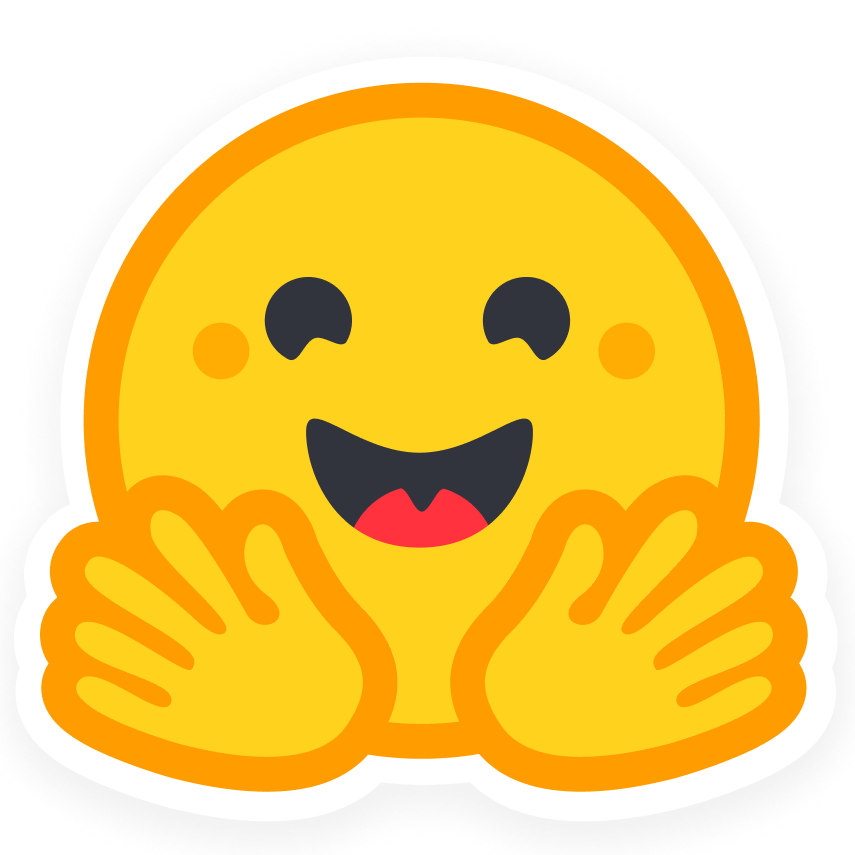}} ~\textbf{\small{Dataset}} & {\textcolor{cyan}{\small \url{https://huggingface.co/datasets/megagonlabs/cypherbench}}} \\
    \raisebox{-1.5pt}{\includegraphics[height=1.05em]{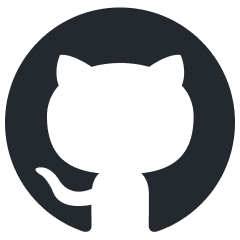}} ~\textbf{\small{Code}} & {\textcolor{cyan}{\small \url{https://github.com/megagonlabs/cypherbench}}}
\end{tabular}
\end{center}
\vspace{-0.9em}
\end{abstract}

\section{Introduction}

Graphs, as a natural modality for modeling entity-relation data, have been widely used for storing both large-scale encyclopedic knowledge and domain-specific enterprise data. Compared to raw textual documents, graphs enable efficient processing of complex multi-hop aggregation
\sajhidden{aggregation might not be the most impactful argument for graphs. Property graphs are definitely more suitable for that. One possible approach can be to start with property graphs right off the bat} 
queries (\eg \textit{What is the average height of point guards who have played for the Toronto Raptors?}), where the answer might depend on information spread across thousands of documents or even the entire corpus. Graphs also provide a more compact representation of knowledge. For example, Wikidata \cite{vrandevcic2014wikidata} contains on average 4.6 times the entities covered by Wikipedia across the domains we experimented with. These advantages has motivated decades of research in knowledge graphs and knowledge base question answering (KBQA) \cite{berant-etal-2013-semantic,moon-etal-2019-opendialkg,dubey2019lc,gu2021beyond,cao-etal-2022-kqa}, as well as the recent proposal of GraphRAG \cite{edge2024local,peng2024graph}.

\begin{figure}[ht]
    \centering
    \includegraphics[width=\linewidth]{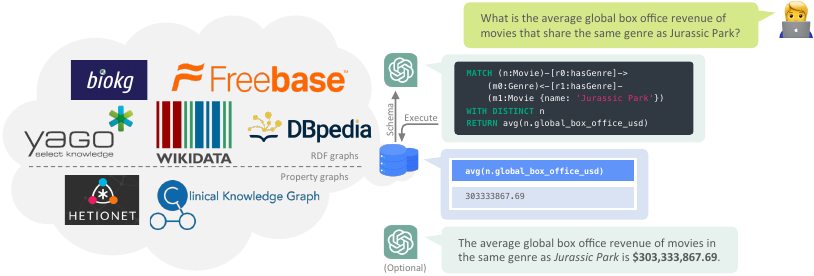}    
    \vspace{-0.1cm}\caption{An illustration of Cypher as a unified interface for retrieval over both RDF and property graphs. A typical graph retrieval or RAG workflow involves:  1) text-to-Cypher translation using an LLM, 2) Cypher query execution, and optionally, 3) final answer generation.}
    \label{fig:text2cypher}
    \vspace{-0.1cm}
\end{figure}

However, retrieval\footnote{Graph retrieval can be considered as a broader task than KBQA, as it is not only essential for question answering but also for other tasks such as fact checking \cite{kim-etal-2023-factkg}.} from modern encyclopedic knowledge graphs \cite{vrandevcic2014wikidata,bollacker2008freebase,suchanek2007yago,lehmann2015dbpedia}, which are predominantly based on RDF, has proven challenging even with the use of LLMs, unlike the success achieved with relational databases using text-to-SQL. Previous studies in KBQA (detailed in \autoref{sec:related}) typically focused on simplified settings for evaluating algorithmic improvements, using either smaller subgraphs \cite{cao-etal-2022-kqa,baek-etal-2023-direct,emonet2024llm}, simple queries without aggregation or grouping \cite{berant-etal-2013-semantic,dubey2019lc,nie-etal-2022-graphq,baek-etal-2023-direct}, or assuming the entity identifiers are provided \cite{furrer2020compositional,kovriguina2023sparqlgen,banerjee2022modern}, which limits their practical application in real-world scenarios. As a result, leading LLM frameworks, including LangChain and LlamaIndex, have only minimal support for retrieval from modern RDF knowledge graphs as of the writing of this paper. Instead, they have prioritized retrieval from \textit{property graphs}, which are typically domain-specific with smaller schemas \cite{santos2022knowledge,himmelstein2017systematic}, using text-to-Cypher generation.

% \footnote{\url{https://www.langchain.com/}}
% \footnote{\url{https://www.llamaindex.ai/}}

We analyze the root cause of why retrieval from RDF knowledge graphs is challenging in \autoref{sec:rdf-to-property} and propose transforming them into multiple smaller property graphs. These property graphs function as domain-specific \textit{views} (analogous to views in relational databases) that can be efficiently queried by LLMs. This approach not only simplifies retrieval from modern RDF knowledge graphs, but also enables the use of Cypher as a unified query language for both RDF graphs and property graph databases like Neo4j that are widely used in enterprise.

As a proof of concept, we introduced \textbf{CypherBench}, a collection of 11 property graphs transformed from Wikidata (detailed in \autoref{sec:graph}). Each graph contains the \textit{complete} set of entities and relations from Wikidata that conform to a domain-specific schema. Together, these graphs include a total of 7 million entities, covering roughly 25\% of Wikipedia and 6\% of Wikidata. In addition, we constructed over 10,000 natural language questions that spans 12 types of graph matching patterns (detailed in \autoref{sec:task}). Notably, we include \textit{global queries} which have been largely overlooked by prior KBQA benchmarks. The benchmark presents significant challenges, with \texttt{gpt-4o} achieving 60.18\% execution accuracy, and no LLMs with <10B parameters surpassing 20\%. 

In summary, this paper makes the following main contributions: 
\begin{itemize}[left=10pt]
    % \item A proposal to use property graphs and Cypher as a unified interface for both RDF and property graph retrieval.
    \item A novel methodology to enable efficient and accurate text-to-Cypher retrieval over modern RDF knowledge graphs.    
    \item A collection of 11 large-scale property graphs with 7 million entities, serving as groundwork for future graph retrieval research and as a high-quality locally-deployable knowledge source (with a knowledge cutoff of April 2024) for augmenting LLMs.
    \item An RDF-to-property-graph transformation engine for Wikidata that creates the aforementioned graphs. It handles triple transformation, datatype conversion, and unit standardization to produce clean, schema-enforced property graphs as output.
    \item A text-to-Cypher / KBQA benchmark with over 10,000 instances spanning 12 types of graph patterns, covering global queries, multi-hop queries, temporal queries and aggregation queries.
    \item An automatic text-to-Cypher task generation pipeline that creates the aforementioned benchmark. It can be used to generate (question, Cypher) pairs for any Neo4j graph database endpoint.
    % \item An automatic template-based text-to-Cypher task generation pipeline that creates the aforementioned benchmark, supports custom templates and works for any Neo4j graph database.
    \item A set of related tools, including graph deployment Docker, evaluation scripts, and graph visualization tools.
\end{itemize}

\section{Knowledge Graph Modeling in the LLM Era}\label{sec:rdf-to-property}

 \subsection{Preliminaries: Knowledge Graphs, RDF and Property Graphs} \label{subsec:preliminary}

The Resource Description Framework (RDF) and property graph are two approaches to modeling and querying knowledge graphs. We begin with an abstract definition of a knowledge graph and then discuss how it is implemented in RDF and property graphs.

In its most basic form, a knowledge graph is a list of relations (also called triples) in the format (subject entity, relation type, object entity), such as ("LeBron James", \texttt{playsFor}, "LA Lakers"). \footnote{Note that relation types are also called properties in Wikidata or predicates in KBQA literature. In this paper, we use "properties" specifically to denote entity and relation attributes in property graphs.} Additionally, entities are often assigned entity types (e.g., \texttt{Person}). Entities and relations can also be associated with literal values as properties (\eg \texttt{receivesAward.year}).

The most popular public knowledge graphs to date (\eg Wikidata, Freebase, and DBpedia) are predominantly based on the RDF and queried using SPARQL. In RDF, entities are stored and accessed using Internationalized Resource Identifiers (IRIs). Entity properties, including entity names, are stored as relations, with the subject being the entity IRI and the object being a literal value. To store relation properties, RDF uses a process called reification, which creates a copy of the relation as a special entity\footnote{Specifically, the statement node in Wikidata and the CVT node in Freebase.} and links it to the relation property using an additional relation.

%\yanlin{Mention other usage of Neo4j, mention datahub} 
Property graph databases have gained significant popularity in industry in recent years, with Neo4j being the most popular graph database management system today\footnote{According to \url{https://db-engines.com/en/ranking/graph+dbms}}. Unlike RDF, the property graph model treats entities and relations as objects, each of which can be assigned types and have associated properties. In property graphs, entities are often accessed directly using their names.

\subsection{Why is retrieval over modern KG hard?}
Retrieval over modern encyclopedic knowledge graphs, which are predominantly RDF graphs (shown in \autoref{fig:text2cypher}), poses significant challenges due to several factors.

\stitle{Overly large schemas.}~~Modern encyclopedic knowledge graphs aim to cover entities and relations across all domains within a single graph, resulting in an extremely large schema that far exceeds the context window sizes of typical LLMs. For instance, Wikidata currently includes over 4 million entity types and 12,000 relation types. Furthermore, RDF graphs allow entities of arbitrary type to serve as subjects or objects for the same relation types, which further increases the number of unique relation schemas. 

%Use of resource identifiers
\stitle{Use of resource identifiers.}~~
SPARQL queries require identifiers for entities, entity types, relation types which must be obtained via external linkers \cite{sakor2020falcon,feng-etal-2023-calibrated}. This also makes SPARQL queries less readable. For instance, consider the SPARQL and Cypher queries for the question ``$Q4.$ What are the names of taxa that feed on Synsphyronus lathrius?'':\newline
    \begin{minipage}[t]{0.47\textwidth}
    \vspace{-5pt}
        \begin{tcolorbox}[title={\scriptsize SPARQL}, boxrule=0.8pt, colframe=gray, boxsep=2pt, left=2pt, right=2pt, top=2pt, bottom=2pt]
            \begin{lstlisting}[style=TinyCypher]
SELECT ?name WHERE {
  item wdt:P31/wdt:P279* wd:Q16521.
  ?item wdt:P1034 wd:Q10687580.
  ?item rdfs:label ?name. FILTER(LANG(?itemLabel) = "en")
}
\end{lstlisting}
        \end{tcolorbox}
    \end{minipage}\hspace{0.01\textwidth}
    \begin{minipage}[t]{0.47\textwidth}
    \vspace{-5pt}
        \begin{tcolorbox}[title={\scriptsize Cypher}, boxrule=0.8pt, colframe=gray, boxsep=2pt, left=2pt, right=2pt, top=2pt, bottom=2pt]
            \begin{lstlisting}[style=TinyCypher]
MATCH (n:Taxon)-[r0:feedsOn]->(m0:Taxon {name: 'Synsphyronus lathrius'})
RETURN n.name
\end{lstlisting}
        \end{tcolorbox}
    \end{minipage}

\stitle{Overlapping relation types.}~~Wikidata contains semantically overlapping relation types that are created for domain-specific usage. For instance, there are at least six relation types to indicate the starting time of an entity: {\small \texttt{start time (P580)}, \texttt{inception (P571)}, \texttt{date of official opening (P1619)}, \texttt{date of first performance (P1191)}, \texttt{publication date (P577)}, \texttt{service entry (P729)}}. This leads to considerable confusion when selecting the correct relation type to use during retrieval. 

\stitle{Lack of normalization.}~~RDF does not enforce type constraints and standardized units on values. As a result, literal values in Wikidata often appear with different units (\eg centimeters and feet for heights) and sometimes incorrect types, which leads to incorrect results when computing aggregation over these values.

\subsection{Hasn't KBQA already solved KG retrieval?}
KBQA requires graph retrieval to answer questions. However, most existing studies focused on simplified settings to evaluate algorithmic improvements. For example, a common simplification made by recent work is assuming that the entity and relation identifiers are already provided \cite{furrer2020compositional,kovriguina2023sparqlgen,banerjee2022modern,luoreasoning,yu2022decaf,jiangunikgqa}, which reduces the task to retrieval over a small local subgraph.  Moreover, many studies use custom-designed intermediate logical forms that lack support for certain graph querying functionalities (\eg relation properties querying, grouping, variable-length path matching) \cite{berant-etal-2013-semantic,berant-liang-2014-semantic,gu2021beyond,yih-etal-2015-semantic,cao-etal-2022-kqa,nie-etal-2022-graphq,li-etal-2023-shot}. As a result, the majority of existing KBQA approaches (see \autoref{sec:related} for a complete categorization) struggle with some or all of the following types of queries: 
1) queries involving relation properties, such as time-sensitive queries; 2) global queries that do not contain any named entities; and 3) queries requiring complex aggregations over a large number of entities.

\subsection{Our Proposal: Property Graphs and Cypher as a Unified Interface}

\begin{figure}[ht!]
    \centering
    \includegraphics[width=\linewidth]{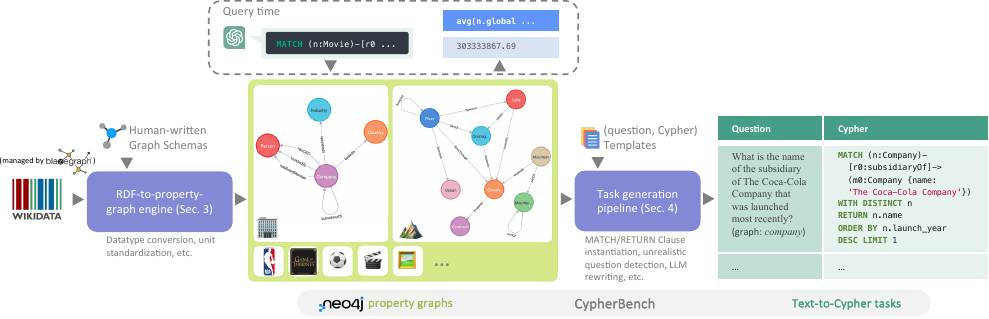}    \vspace{-0.1cm}\caption{CypherBench construction process: Wikidata is transformed into schema-enforced property graphs, which enables efficient and accurate text-to-Cypher querying. These property graphs are then used to generate text-to-Cypher tasks.}
    \label{fig:workflow}\vspace{-0.1cm}
\end{figure}

To address the aforementioned challenges, we propose transforming the RDF graph into multiple domain-specific \textit{property graphs} and using Cypher to query them (see \autoref{fig:workflow} for an illustration). We chose Cypher, the query language for Neo4j, because of its widespread adoption in open-source projects (including LLM frameworks and GraphRAG).  These property graphs function as (materialized) \textit{views} on top of the original RDF graph that can be queried efficiently by LLM. Each property graph view contains all data that conforms to its respective schema and can be updated when the underlying RDF data changes. These views operate independently, may overlap, and may be created on-demand, offering substantial flexibility. 

This design enables scaling to a large number of domains without introducing an overly complex schema and eliminates ambiguity from overlapping relation types by contextualizing them within specific domains. Our RDF-to-property-graph transformation layer manages datatype conversion and unit standardization, producing schema-enforced, strictly typed data to ensure the result correctness for aggregation queries. This transformation is also efficient and takes only a few seconds for small graphs (fewer than 10,000 entities), as it is is achieved by executing SPARQL queries and aggregating the results, rather than processing the entire RDF dump. 

In the following sections, we instantiated this idea on Wikidata, the largest knowledge graph today. We demonstrate that a direct prompting baseline using \texttt{gpt-4o}, without the use of external linkers or retrievers, achieves reasonable performance.

\section{Transforming RDF to Property Graphs} \label{sec:graph}

In this section, we introduce our approach for transforming RDF, specifically Wikidata, into property graphs as the initial step in building CypherBench (\autoref{fig:workflow}). We selected Wikidata because it is the largest and most actively maintained knowledge graph, comprising 114 million entities and having received 270 million edits from over 42,000 active editors in the past 12 months\footnote{\url{https://stats.wikimedia.org/\#/wikidata.org}}.

% \yanlin{Overview subsection? compare with previous RDF-to-property-graph conversion efforst. mention ours do not process raw RDF dump. Relate to the concept of view in databse}

% \subsection{Overview}

% This section describes our process for transforming Wikidata into property graphs. At a high level, we start by selecting a domain and manually curating the property graph schema, along with the mapping from the Wikidata schema to this schema. Next, our RDF-to-property-graph transformation engine uses the mapping to automatically construct a property graph by issuing SPARQL queries to the Wikidata endpoint. The property graph is stored in a DBMS-independent JSON format and is ultimately loaded into Neo4j.

\subsection{Domain-specific Schema Curation}

% \subsection{Schema Curation}
We begin by selecting a domain and manually curating the property graph schema, along with the mapping from the Wikidata schema to this schema. This process typically involves identifying the entity and relation types and their properties relevant to the domain, followed by exploring Wikidata to find the corresponding Wikidata identifiers (QIDs for entity types and PIDs for relation types and properties). Sample entity and relation schemas are shown below:

\begin{minipage}[t]{0.47\textwidth}
    \vspace{-5pt}
        \begin{tcolorbox}[title={\scriptsize Sample Entity Schema with Wikidata Mappings}, boxrule=0.8pt, colframe=gray, boxsep=2pt, left=2pt, right=2pt, top=2pt, bottom=2pt]
            \begin{lstlisting}[style=TinyJSON]
{
  "label": "Movie",
  "wd_source": "Q11424",
  "properties": [
    {
      "label": "runtime_minute",
      "wd_source": "P2047",
      "datatype": "float",
      "quantity_unit": "minute",
      "quantity_convert_unit": true
    }
  ]
}
\end{lstlisting}
        \end{tcolorbox}
    \end{minipage}\hspace{0.01\textwidth}
    \begin{minipage}[t]{0.47\textwidth}
    \vspace{-5pt}
        \begin{tcolorbox}[title={\scriptsize Sample Relation Schema with Wikidata Mappings}, boxrule=0.8pt, colframe=gray, boxsep=2pt, left=2pt, right=2pt, top=2pt, bottom=2pt]
            \begin{lstlisting}[style=TinyJSON]
{
  "label": "receivesAward",
  "wd_source": "P166",
  "subj_label": "Movie",
  "obj_label": "Award",
  "properties": [
    {
      "label": "year",
      "wd_source": "P585",
      "datatype": "int"
    }
  ]
}
\end{lstlisting}
        \end{tcolorbox}
    \end{minipage} 

% \begin{table*}[ht]
% \scriptsize 
% % \vspace{-5pt}
% \centering
% \begin{tabular}{p{3.5cm} p{5cm}}
% \toprule
%  \textbf{Graph Schema} & \textbf{Entity/Relation Properties} \\
% \midrule
% \raisebox{-0.9 \height}{\includegraphics[width=3.5cm]{figures/graphs/company_schema.png}} &
% \RaggedRight \tiny
% \nlabel{Company}  \nprop{name \textcolor{white}{STR}}  \nprop{launch\_year \textcolor{white}{INT}} \quad  \nlabel{Country}  \nprop{name \textcolor{white}{STR}} \quad  \nlabel{Person}  \nprop{place\_of\_birth \textcolor{white}{STR}}  \nprop{name \textcolor{white}{STR}}  \nprop{gender \textcolor{white}{STR}}  \nprop{date\_of\_death \textcolor{white}{DATE}}  \nprop{date\_of\_birth \textcolor{white}{DATE}}  \nprop{country\_of\_citizenship \textcolor{white}{LIST[STR]}} \quad  \nlabel{Industry}  \nprop{name \textcolor{white}{STR}} \newline   \rlabel{hasBoardMember}  \rprop{start\_year \textcolor{white}{INT}}  \rprop{end\_year \textcolor{white}{INT}} \quad  \rlabel{hasCEO}  \rprop{start\_year \textcolor{white}{INT}}  \rprop{end\_year \textcolor{white}{INT}}
% \\
% \bottomrule
% \end{tabular}
% \caption{ Schema of the \textit{company} graph. The color of the property boxes indicates whether they are \textcolor{beige}{entity properties} \big(\eg \raisebox{6pt}{\scriptsize \nprop{name \textcolor{white}{STR}}}\big) or \textcolor{olive}{relation properties} \big(\eg \raisebox{6pt}{\scriptsize \rprop{start\_year \textcolor{white}{INT}}}\big). The schemas of all 11 graphs are displayed in \autoref{tab:all_graphs}. }
% \label{tab:sample_graph}
% \end{table*}

% \begin{figure}[h]
\begin{wrapfigure}{r}{0.6\textwidth}
\vspace{-0.2cm}
    \centering
    \includegraphics[width=0.9\linewidth]{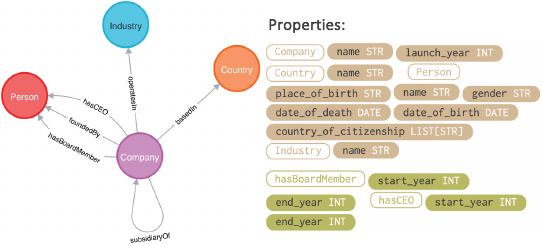} \vspace{-0.2cm}   \caption{ Schema of the \textit{company} graph with \textcolor{beige}{entity properties} an \textcolor{olive}{relation properties}. See \aref{appendix:cypherbench} for other graphs.}
    \label{fig:sample_graph}
\end{wrapfigure}
% \end{figure}

Each property is assigned a datatype. Properties that represent quantities are also given a unit, which is indicated in the property label (e.g., \texttt{runtime\_minute}) to inform LLMs during graph retrieval.

The authors created all 11 property graph schemas from scratch, with an average time investment of approximately 4 hours per graph.  The complete schema of a sample graph is presented in \autoref{fig:sample_graph}, and the schemas for all graphs are listed in \aref{appendix:cypherbench}.

\subsection{Automatic RDF-to-property-graph Transformation}

% \saj{how are we different from GraphQ-IR~\cite{nie-etal-2022-graphq}?}

% \saj{also we should have some discussion on selecting Wikidata over DBpedia or other candidates.}

Next, our RDF-to-property-graph engine issues SPARQL queries to Wikidata, using the identifiers from the curated schema 
to fetch all entities and relations that conform to the schema. For example, the following SPARQL query (simplified for illustration) fetches all Wikidata \texttt{award received (P166)} relations where the subject is an instance of \texttt{Film (Q11424)} and the object is an instance of \texttt{Award (Q618779)}. These relations are then converted into \texttt{receivesAward} relations in the target property graph. 

%\saj{this can potentially be part of the process diagram}
\begin{minipage}[t]{\textwidth}
\vspace{-5pt}
    \begin{tcolorbox}[title={\scriptsize Wikidata SPARQL}, boxrule=0.8pt, colframe=gray, boxsep=2pt, left=2pt, right=2pt, top=2pt, bottom=2pt]
        \begin{lstlisting}[style=TinyCypher]
SELECT DISTINCT ?subj ?obj ?statement
WHERE {
  ?subj wdt:P31 wd:Q11424. ?obj wdt:P31 wd:Q618779. ?subj p:P166 ?statement. ?statement ps:P166 ?obj.
}
\end{lstlisting}
    \end{tcolorbox}
\end{minipage} 

The actual conversion engine further incorporates the following functionalities:
\sajhidden{Consider a boolean property \emph{marital status} of a \emph{Person} in a graph. The property value can be expressed using a boolean (True/False) or integer (0/1) data type. The absence of any ``type constraint'' may lead to incorrect query generation given a natural language question. For example, when filtering out \emph{Person} by their marital status or when performing aggregation over married or unmarried personnel (\eg average salary.) Similar observations are reported in text-to-SQL literature~\cite{bird-bench} where noisy or missing data type information in schema degrades SQL generation performance.}
 
\stitle{Datatype conversion.}~~The engine enforces type constraints on property values by converting them into one of the following types and discarding those that cannot be converted: \texttt{str}, \texttt{int}, \texttt{float}, \texttt{date}, \texttt{list[str]}.

\stitle{Date conversion.}~~Wikidata stores precision for time values ranging from seconds up to centuries or more (\eg a historical event might be recorded with century-level precision like 18th century). For date properties, we retrieve the precision using the predicate \texttt{wikibase:timePrecision} and keep only those with a precision at least as fine as a date, which are ultimately mapped to Neo4j's \texttt{Date} values.
\sajhidden{Human-curated knowledge graphs may contain semantically overlapping relation types that are created for different domains and often used inconsistently.}

\stitle{Unit standardization.} ~~The engine enforces standardized units (\eg centimeters) on property values that represent quantities by converting all values that can be converted. This eliminates the need for unit conversion during retrieval and ensures accuracy of aggregation queries.

% \saj{As data and knowledge may change over time, knowledge graphs maintain specialized fields or identifiers to track temporally evolving information.}
\stitle{Rank filtering.}~~Wikidata uses rank (\texttt{wikibase:rank}) to indicate the reliability or recency of a relation, which can be one of three values: preferred, normal, or deprecated. For time-sensitive relations (\eg the president of the United States), the currently valid entries are typically marked as preferred, while previously valid entries are marked as normal with additional properties indicating the relevant time period. For time-sensitive relations with time qualifiers in the target schema, we fetch all Wikidata relations with non-deprecated ranks, along with their starting and ending times. For other types of relations, we fetch only the highest-ranked available relation. 

\stitle{Selective entity fetching.}~~The engine supports fetching only entities linked to certain relations to avoid out-of-memory issues for very broad entity types (e.g., people or organizations). For instance, in the movie graph, instead of fetching all instances of \texttt{Person} from Wikidata, the engine limits the fetch to only those connected to a \texttt{Movie} through relations like \texttt{directedBy} or \texttt{hasCastMember}.

%\saj{can this be a general step not only for wikidata but other DBs?}

% \begin{itemize}[left=10pt]
%     \item 
   
% \end{itemize}

% \stitle{Lack of type constraints}~~RDF does not enforce type constraints on literal values of the same property. \saj{Consider a boolean property \emph{marital status} of a \emph{Person} in a graph. The property value can be expressed using a boolean (True/False) or integer (0/1) data type. The absence of any ``type constraint'' may lead to incorrect query generation given a natural language question. For example, when filtering out \emph{Person} by their marital status or when performing aggregation over married or unmarried personnel (\eg average salary.)} Similar observations are reported in text-to-SQL literature~\cite{bird-bench} where noisy or missing data type information in schema degrades SQL generation performance.

% \stitle{Overlapping and ambiguous relation types.}~~  \saj{this seems like schema ambiguity resulting from redundancy. but this doesn't seem like a general issue to KGs rather an artifact of the KG developer. This should go to S3.2}

The SPARQL queries issued by the engine are executed against a local Wikidata endpoint loaded with the April 2024 Wikidata dump, allowing us to bypass the time limit of the public endpoint, and the results are aggregated into the final property graph. The transformation time ranges from seconds to hours, depending on the graph size. The graph statistics are shown in \autoref{tab:graph_stats}. The property graph is stored in a DBMS-independent JSON format and ultimately deployed using a custom Neo4j Docker image that initializes the data from the JSON file upon startup.
\sajhidden{move to appendix for ARR. We can simply say in the experiment section that, our data collection was done on the April 2024 Wikidata dump.}

% \subsection{Neo4j Deployment} \label{subsec:neo4j_deployment}

 % \sajhidden{move to appendix for ARR}

\section{Constructing Questions} \label{sec:task}

\subsection{Graph Retrieval via Text-to-Cypher}

With 11 large-scale property graphs as a testbed, the next step is to construct questions that require graph retrieval to be answered. The most straightforward approach for graph retrieval is to translate the question into a graph database query (\eg Cypher) that fetches the relevant information or the answer. Alternative approaches used in previous KBQA studies (see \autoref{tab:graph_retrieval_methods} for an overview of existing graph retrieval methods), such as top-$k$ embedding-based retrieval, typically cannot handle complex aggregation queries where the answer may depend on thousands of entities. While we adopt text-to-Cypher as the primary graph retrieval approach and develop a benchmark consisting of (question, Cypher) pairs, we also record execution results of the Cypher queries as answers so that it can serve as a generic KBQA benchmark for evaluating non-Cypher-based approaches.  

A text-to-Cypher task can be formulated as follows: given the graph schema\footnote{The graph schema is usually obtained by executing a special Cypher query against the database endpoint and can be cached for future use (see \aref{appendix:schema} for details).} and a natural language question as input, the goal is to output an executable Cypher query that returns the desired answer. We require the Cypher query to produce the final answer on its own, thus eliminating the need for an additional answer generation step with an LLM as in a standard RAG pipeline.

Our Text-to-Cypher task generation pipeline involves two main steps: 1) generating initial (question, Cypher) pairs with diverse graph patterns using templates, and 2) rewriting the questions to sound more natural using a LLM.

\begin{table*}[h]
\scriptsize 
\vspace{-5pt}
\centering
\begin{tabular}{p{1.78cm} p{4.97cm} p{5.95cm}}
\toprule
\textbf{Pattern/Template} & \textbf{Sample Question} & \textbf{Cypher Query} \\
\midrule
\raisebox{-0.6\height}{\includegraphics[height=0.4cm]{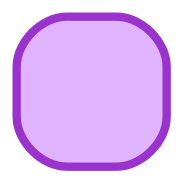}}\newline {\tiny \texttt{(Basic MATCH)}} & $Q1.$ What are the names of terrorist attacks that occurred before March 13th, 1997? \hfill (\textit{terrorist~attack})  & \vspace{-1.6ex}\begin{lstlisting}[style=TinyCypher]
MATCH (n:TerroristAttack) WITH DISTINCT n
WHERE n.date < date('1997-03-13') RETURN n.name
\end{lstlisting}\vspace{-1.6ex}
\\
\midrule
\raisebox{-0.6\height}{\includegraphics[height=0.4cm]{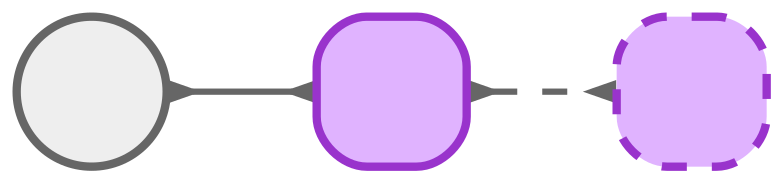}} \newline\newline\texttt{Optional Match}\newline {\tiny \texttt{(Special MATCH)}} & $Q10.$ Provide the names of all aircraft models manufactured by ATR, along with the number of flight accidents each has been involved in. \hfill (\textit{flight~accident}) & 
\vspace{-1.6ex}\begin{lstlisting}[style=TinyCypher]
MATCH (n:AircraftModel)-[r1:manufacturedBy]->
    (m1:AircraftManufacturer {name: 'ATR'}) 
OPTIONAL MATCH (n:AircraftModel)<-
        [r0:involves]-(m0:FlightAccident)
WITH n, count(DISTINCT m0) AS num
RETURN n.name, num
\end{lstlisting}\vspace{-1.6ex}
\\
\midrule
\texttt{AGGREGATE}\newline {\tiny \texttt{(RETURN Template)}} & $Q18.$ What is the average longest lifespan of taxa that feed on Leporidae? \hfill (\textit{biology})  & \vspace{-1.6ex}\begin{lstlisting}[style=TinyCypher]
MATCH (n:Taxon)-[r0:feedsOn]->(m0:Taxon {name: 'Leporidae'}) WITH DISTINCT n
RETURN avg(n.longest_lifespan_years)
\end{lstlisting}\vspace{-1.6ex}
\\
\bottomrule
\end{tabular}
\caption{ A basic MATCH pattern, a special MATCH pattern, and a RETURN template, along with sample questions. The \textcolor[HTML]{9933cc}{nodes in purple} denote the answer entities. Square nodes \big(\raisebox{-0.1cm}{\includegraphics[height=0.4cm]{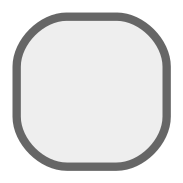}\includegraphics[height=0.4cm]{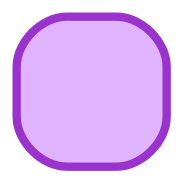}}\big) denote all entities of a particular type, while circular nodes \big(\raisebox{-0.1cm}{\includegraphics[height=0.4cm]{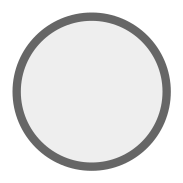}\includegraphics[height=0.4cm]{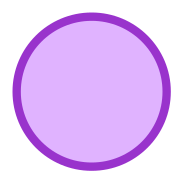}}\big) represent named entities. Nodes and edges with dashed lines \big(\raisebox{-0.1cm}{\includegraphics[height=0.4cm]{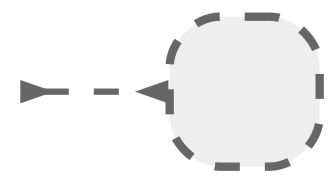}}\big) are optional. The complete list of patterns are provided in \autoref{tab:graph_patterns} and \autoref{tab:return_patterns}.}
\label{tab:sample_templates}
\end{table*}

\subsection{Preliminaries: Cypher Query Structure}

A Cypher query typically begins with a \texttt{MATCH} clause, which identifies the subgraphs that match the specified graph pattern. Following this, the remaining Cypher clauses perform various transformations---such as filtering, ranking, or aggregation---to generate the desired result. For simplicity, we refer to all clauses that follow the \texttt{MATCH} clause as the \texttt{RETURN} clause, which may include \texttt{WHERE} , \texttt{WITH}, \texttt{ORDER BY} and \texttt{RETURN} clauses.

% This can be followed by various transformations (\eg sorting with \texttt{ORDER BY}) and concludes with a final \texttt{RETURN} clause. For simplicity, we refer to all the clauses following the \texttt{MATCH} clause as the \texttt{RETURN} clause.

\subsection{Graph Pattern Design} \label{subsec:pattern_design}

At the core of graph retrieval is the task of locating the subgraph relevant to the query, which is a fundamental feature of all mainstream graph database query languages (\eg \texttt{MATCH} clauses in Cypher, \texttt{WHERE} clauses in SPARQL, etc.). To ensure a balanced distribution across various graph matching patterns, we adopt a template-based generation approach rather than crowd-sourcing (as shown in \autoref{tab:benchmark_compare}, even crowd-sourced multi-hop QA benchmarks like HotpotQA tend to be biased toward only a few types of graph matching patterns). 
% \sajc{we need to define graph reasoning patterns.}

% \yanlin{Mention somewhere that existing KBQA benchmarks are mostly what is the X of Y or what is the X of Y of Z?}

% Each node generally corresponds to a node variable in Cypher or SPARQL.
Graph patterns can be categorized based on the isomorphism structure of an undirected graph (see \autoref{tab:sample_templates} for sample patterns and \autoref{tab:graph_patterns} for the complete notations).  As shown in \autoref{tab:graph_patterns}, we define seven basic graph patterns, covering all possible isomorphism structures with a single answer node and up to two edges. Additionally, we design five special graph patterns that cover comparison, grouping, optional matching, time-sensitive queries, and union. 

% \sajhidden{this paragraph suddenly appears.}

We compare the graph patterns covered by representative benchmark in \autoref{tab:benchmark_compare}. A notable observation is that most existing KBQA benchmarks overlook \textit{global queries} that GraphRAG targets \cite{edge2024local}, which we define as queries without any specific named entities. These can range from simple listing queries like ``$Q13.$ List the names of all teams'' \big(\raisebox{-0.1cm}{\includegraphics[height=0.4cm]{figures/match_patterns/basic_1.png}}\big) to more complex ones like ``$Q7.$ What are the unique countries of citizenship of individuals who both wrote and acted in the same movie?'' \big(\raisebox{-0.1cm}{\includegraphics[height=0.4cm]{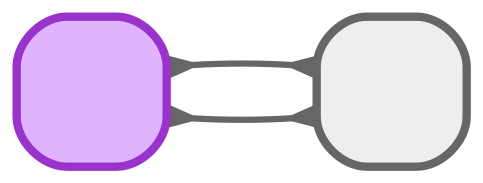}}\big). The answers to these global queries typically depend on a large number of documents and cannot be easily handled by standard RAG approaches. \sajhidden{this seems like a good motivating example for the paper and can go intro or section 2.}

% The graphical notations are shown in . Purple nodes indicate the answer entities. Square nodes denote all entities of a specific type, while circular nodes represent entities with names.

\subsection{Text-to-Cypher Task Generation}

\subsubsection{\texttt{MATCH} Clause Instantiation}

For each graph pattern, we create multiple Cypher \texttt{MATCH} clause templates by enumerating all possible edge directions, with each \texttt{MATCH} template paired with a human-written question template. For example, one of the (question, Cypher) template for pattern \raisebox{-0.1cm}{\includegraphics[height=0.4cm]{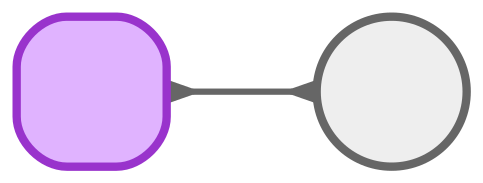}} is \big(\texttt{\small ``\textcolor{mygreen}{\textbf{MATCH}} (n)-[r0]->(m0<name>)''}, \texttt{\small ``\$\{n\_LABEL\} that \$\{r0\_LABEL\} \$\{m0\_name\}''}\big). Next, the template is instantiated by sampling entity types for node variables (\texttt{n}, \texttt{m0}), relation types for edge variables (\texttt{r0}), and entity names for named nodes (\texttt{m0}). We accomplish this by executing a special Cypher query on a sampled subgraph, which ensures that the instantiated \texttt{MATCH} clause returns non-empty results. \sajhidden{this section needs a bit more clairification as we haven't introduced the patterns yet.} 

\subsubsection{\texttt{RETURN} Clause Instantiation}
% \protect\footnote{The \texttt{RETURN} clause here also includes associated clauses, such as \texttt{WHERE} and ORDER BY.}

% After the \texttt{MATCH} clause locates the relevant subgraph, the remaining Cypher clauses apply various transformations—such as filtering, ranking, or aggregation—to produce the desired answer. For simplicity, we refer to this part as the \texttt{RETURN} clause. \sajhidden{we need to define the construct of a Cypher query earlier in Section 2.}

Each instantiated \texttt{MATCH} clause in the special categories is paired with its dedicated \texttt{RETURN} clause template, while those in the basic categories are paired with one of the six templates (showns in \autoref{tab:return_patterns}) that covers basic property fetching  (\texttt{NAME}, \texttt{PROPERTY}), ranking (\texttt{SORT}), filtering (\texttt{WHERE}) and aggregation (\texttt{AGGREGATE}, \texttt{ARGMAX}). The \texttt{RETURN} clause template is instantiated by sampling properties, ranking orders, comparison operators (\eg $\leq$, $\neq$) and aggregate functions (\eg \texttt{min}, \texttt{avg}, \texttt{count}). We take the datatype into account when sampling operators and aggregate functions to ensure the question is realistic. For instance, we do not allow $\leq$ on string properties or \texttt{avg} on dates. Similar to \texttt{MATCH} clauses, each \texttt{RETURN} clause also has a textual template that is instantiated and combined with the one for the \texttt{MATCH} clause to form the complete question.

One subtle design choice we made is to ensure the Cypher always returns literal values, such as entity names, instead of node objects. This allows the benchmark to be used to evaluate non-Cypher graph retrieval methods in the future.

% \begin{table*}[h]
\begin{wraptable}{r}{0.55\textwidth}
\scriptsize 
\centering
\begin{tabular}{l p{4cm} r}
\toprule
\textbf{Split} & \textbf{Graphs} & \textbf{\#question} \\
\midrule
Train & \textit{art, biology, soccer, terrorist~attack} & 8817 \\
\midrule
Test & \textit{company, fictional character, flight~accident, geography, movie, nba, politics} & 2488 \\
% \midrule
% Total & & 11305 \\
\bottomrule
\end{tabular}
\caption{ Statistics of the data splits. }
\label{tab:question_stats}
\end{wraptable}
% \end{table*}

\begin{figure}[H]
    \centering
    \vspace{-0.1cm}\includegraphics[width=\linewidth]{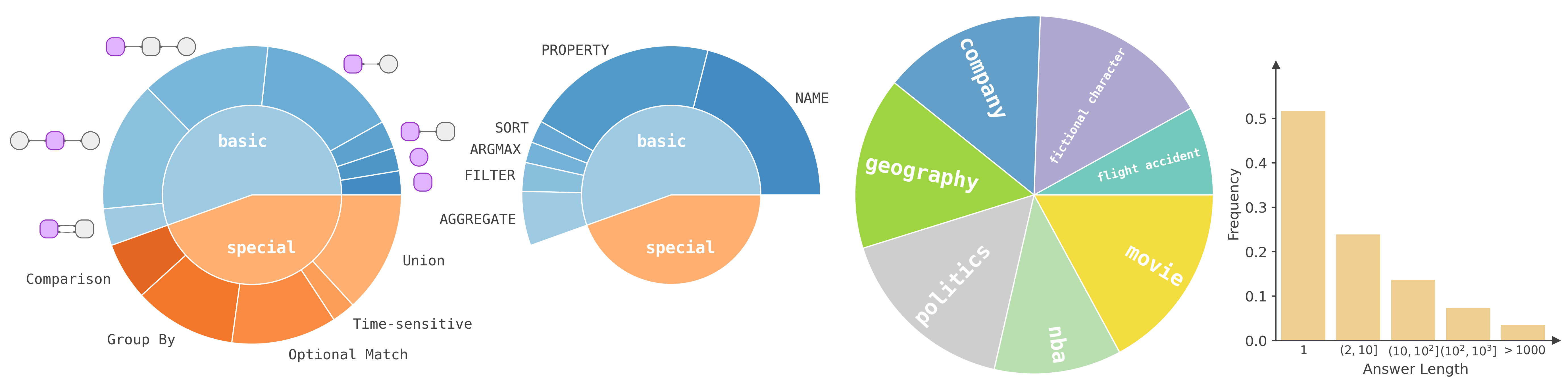}   \vspace{-0.4cm}\caption{Distribution of graph matching patterns, \texttt{RETURN} templates, domains, and answer lengths (number of rows in the answer) in the CypherBench test set.}
    \label{fig:question_distribution}\vspace{-0.1cm}
\end{figure}

The questions are split into training and test sets by domain, with 4 graphs allocated for training and 7 graphs for testing (\autoref{tab:question_stats}). We remove Cypher queries that produce more than $10^5$ records or take more than 30 seconds to execute. The distribution of the test set across four dimensions, as shown in \autoref{fig:question_distribution}, demonstrates that the benchmark is both diverse and balanced. 

% Notably, it also includes a significant portion of questions where the answers contain over 1,000 records.

\subsubsection{Detecting Semantically Unrealistic Questions}

% \footnote{Cypher: \texttt{\small \textcolor{mygreen}{\textbf{MATCH}} (n:Character)<-[r0:hasSpouse]-(m0:Character)-[r1:hasSpouse]->(m1:Character \{name: 'Rhaenyra Targaryen'\}) \textcolor{mygreen}{\textbf{RETURN}} n.name}.} 
Blind sampling often results in semantically unrealistic or uninteresting questions, such as ``Who is married to someone married to Rhaenyra Targaryen?'' This issue has been noted in previous KBQA studies \cite{su-etal-2016-generating}, where it was addressed using heuristics that might incorrectly exclude realistic questions. In this work, we take a more systematic approach by modeling the cardinality and participation characteristics of relationships: \sajhidden{this is an interesting contribution. but are these the only conditions to test of realisticness?}

\begin{itemize}[left=10pt]
    \item \textit{Cardinality.} A directional relationship\footnote{A relationship here can be considered as an edge in the graph schema—a (subject entity type, relation type, object entity type) triplet.} can have one of four cardinalities: one-to-one, one-to-many, many-to-one, or many-to-many. For example, \texttt{hasSpouse} represents a one-to-one relationship. Cardinality is used to detect unrealistic groupings—groupings that would always result in a single member per group (\eg ``For each character, return the number of fathers.''), as well as unnecessary consecutive inverse relations, as in the \texttt{hasSpouse} example.

    \item \textit{Participation.} The participation of the subject or object in a relationship describes whether its entity instances are always associated with that relationship.  Participation can either be total (\eg \texttt{Movie} in \texttt{directedBy}, assuming every movie has a director) or partial (\eg \texttt{Movie} in \texttt{receivesAward}, as not all movies receive awards). We use participation to detect redundant conditions (\eg “List all movies directed by someone”). 

    \item \textit{Entailment.}  A relationship can imply another (e.g., \texttt{hasFather} implies \texttt{hasParent}). This information is also used to detect redundant conditions.
\end{itemize}
                         
The characteristics of cardinality, participation, and entailment are manually determined rather than derived from the data, due to the presence of missing data. They are also not enforced as constraints on the data for the same reason.

\subsection{Question Rewriting and Verification}

We employ LLMs to rewrite the template-generated questions into more natural-sounding questions and to diversify their phrasing. However, we intentionally preserve entity names and string values to avoid introducing the additional complexity of entity linking. This design choice allows us to evaluate LLMs directly through prompting without introducing external linkers or retrievers, leaving the task of entity linking for future work. 

We observe that LLMs sometimes alter the meaning of the question during rewriting. For example,
a common mistake is reversing the direction of relations (\eg confusing ``companies that are
subsidiaries'' with ``companies that have subsidiaries''). To address this, we implement three rounds
of verification and revision using LLMs. Finally, the authors inspect all instances in the test set to ensure they are correct.

% The template-generated questions are often ungrammatical or unnatural. Therefore, we employ LLMs to rewrite them into more natural-sounding questions and to diversify their phrasing. However, we intentionally preserve entity names and string values to avoid introducing the additional complexity of entity linking. This design choice allows us to evaluate LLMs directly through prompting without introducing external linkers or retrievers, leaving the task of entity linking for future work.
% \sajhidden{since we are already testing mostly LLMs and since it is a benchmark. we can potentially pick a subset of the questions and introduce schema-linking chanllenges in those questions.}

% allow a database-expert-level LLM to achieve perfect performance independently

% We observe that LLMs sometimes alter the meaning of the question during rewriting. For example, a common mistake is reversing the direction of relations (\eg confusing ``companies that are subsidiaries'' with ``companies that have subsidiaries''). To address this, we implement three rounds of verification and revision using LLMs. Finally, the authors inspect all instances in the test set to ensure they are correct. \sajhidden{let's revisit this. are we doing some reverse translation to verify?}

% \yanlin{Dataset Statistics subsection?}

\section{Evaluation Metrics}

\subsection{Execution Accuracy (EX)}

% We use execution accuracy as our primary evaluation metric. 
Execution accuracy, the standard metric in text-to-SQL evaluation, measures whether the results returned by the predicted query match those returned by the ground truth query. Cypher returns results in a tabular format, thus allowing us to borrow the execution accuracy implementation from the text-to-SQL literature. In this work, we adapt the execution accuracy implementation\footnote{\url{https://github.com/taoyds/test-suite-sql-eval}} from the Spider \cite{yu-etal-2018-spider} leaderboard, which considers two tables as identical if one can be transformed into the other through row and column permutations. \footnote{One difference between Cypher and SQL is that Cypher supports objects (\eg lists and maps) in query results. We serialize these objects to enable direct comparisons.}
\begin{equation}
    \text{EX}(q, \hat{q}) = \mathbbm{1}_{V = \hat{V}}(V, \hat{V})
\end{equation}
where $V$ and $\hat{V}$ are the execution results of the ground-truth and predicted Cypher. The final dataset-level metric is obtained by averaging across all instances. Note that execution accuracy can also be applied to non-Cypher-based graph retrieval approaches in future research, as long as the approach returns results in the tabular format.

\subsection{Provenance Subgraph Jaccard Similarity (PSJS)}

As discussed in \autoref{subsec:pattern_design}, the core task of graph retrieval is to locate the relevant subgraph using the \texttt{MATCH} clause. While a LLM might generate the correct \texttt{MATCH} clause, it can make subtle mistakes such as returning node objects instead of entity names, or including an extra column that was not requested by the question. In other cases, the \texttt{MATCH} clause might be partially correct, either missing or including a few extra entities. All these scenarios would result in zero execution accuracy. 

% \yanlin{PSJS in some cases is stricter than EX - when two Cyphers produce the same answer through different graph matching}
We propose Provenance Subgraph Jaccard Similarity (PSJS)  as an isolated measure of the subgraph matching performance. We define the \textit{provenance subgraph} as the subgraph matched by the \texttt{MATCH} clause, which can be obtained by pairing the \texttt{MATCH} clause with \texttt{RETURN *}. For example, the provenance subgraph for “$Q18.$ What is the average longest lifespan of taxa that feed on Leporidae?” would include the entity Leporidae and all taxa that feed on Leporidae. PSJS is then calculated as the Jaccard similarity --- a standard metric for comparing two sets --- between the provenance subgraph of the predicted Cypher and that of the ground truth Cypher:
% \[
%   \text{PSJS} = \frac{\text{ProvSubgraph}(q)\cap \text{ProvSubgraph}(\hat{q})}{\text{ProvSubgraph}(q)\cup \text{ProvSubgraph}(\hat{q})}
% \]
\begin{equation}
  \text{PSJS}(q, \hat{q}) = \frac{{G}\cap \hat{{G}}}{{G}\cup \hat{{G}}}
\end{equation}
where $G$ and $\hat{G}$ are the provenance subgraphs of the ground-truth and predicted Cypher.

As another example, a predicted Cypher query that satisfies only one condition in a \texttt{UNION} query would receive an execution accuracy of 0 and a PSJS score equal to the fraction of correctly retrieved nodes.

\section{Experiments}

\subsection{Evaluation Details}

We deployed the graphs using a custom Neo4j Docker image \footnote{\url{https://hub.docker.com/repository/docker/megagonlabs/neo4j-with-loader}} on a local server with 1TB memory. Since the Neo4j community edition does not support multiple databases, we ran a separate Docker container for each graph.

To evaluate the zero-shot text-to-Cypher performance of state-of-the-art LLMs, we run a variety of popular LLMs of different sizes on the CypherBench test set\footnote{The training set was not used in this study.}. For each task instance, the model was prompted with the question, the graph schema, and a brief instruction. The complete prompt is shown in \aref{appendix:text2cypher_prompt}. The open-source models and \texttt{yi-large} were run using the Fireworks AI API, \texttt{gemini1.5} and \texttt{claude3.5-sonnet} were run on Google Cloud Vertex AI, while \texttt{gpt-} models were run using OpenAI's API. The cost per run is \$5.5 for \texttt{gpt-4o} and \$0.3 for \texttt{gpt-4o-mini}.

Finally, the predicted Cypher queries were executed on Neo4j using 8-thread parallelization with a 120-second timeout (4x the maximum execution time of the ground-truth Cypher) to compute the metrics.

% \begin{table*}[ht]
\begin{wraptable}{r}{0.55\textwidth}
\scriptsize
\centering
\begin{tabular}{lccc}
\toprule
\textbf{Model} & \textbf{EX (\%)} & \textbf{PSJS (\%)} & \textbf{Exec. (\%)} \\
\midrule
\rowcolor[gray]{0.9}
\multicolumn{4}{l}{\textit{Open-source LLMs (<10B)}} \vspace{2pt} \\
\texttt{llama3.2-3b} & 11.20 & 17.33 & 86.46 \\
\texttt{llama3.1-8b} & 18.82 & 30.98 & 90.67 \\
\texttt{gemma2-9b} & 18.61 & 30.67 & 68.57 \\
\midrule
\rowcolor[gray]{0.9}
\multicolumn{4}{l}{\textit{Open-source LLMs (10-100B)}}
\vspace{2pt} \\
\texttt{mixtral-8x7b} & 19.21 & 37.01 & 59.33 \\
\texttt{qwen2.5-72b} & 41.87 & 56.39 & 86.84 \\
\texttt{llama3.1-70b} & 38.84 & 54.79 & 92.25 \\
\midrule
\rowcolor[gray]{0.9}
\multicolumn{4}{l}{\textit{Proprietary LLMs}}
\vspace{2pt} \\
\texttt{yi-large} & 33.82 & 47.21 & 83.52 \\
\texttt{gemini1.5-flash-001} & 25.26 & 41.46 & 83.65 \\
\texttt{gemini1.5-pro-001} & 39.95 & 57.70 & 86.03 \\
\texttt{gpt-4o-mini-20240718} & 31.43 & 45.91 & 87.39 \\
\texttt{gpt-4o-20240806} & 60.18 & 76.87 & 94.93 \\
\texttt{claude3.5-sonnet-20240620} & \textbf{61.58} & \textbf{80.85} & \textbf{96.34} \\
\bottomrule
\end{tabular}
\caption{Zero-shot execution accuracy (EX), provenance subgraph jaccard similarity (PSJS) and executable percentage (Exec.) on the CypherBench test set.}
\label{tab:main_result}
\end{wraptable}
% \end{table*}

\subsection{Main Results}

As shown in \autoref{tab:main_result}, the best-performing model \texttt{claude3.5-sonnet} achieves an execution accuracy of 61.58\% and a PSJS of 80.85\%, with \texttt{gpt-4o} performing slightly worse. The highest-performing open-source model reaches only 41.87\% execution accuracy, while smaller models in the <10B parameter range achieve less than 20\% execution accuracy. These results highlight the difficulty of CypherBench.

Furthermore, the low PSJS scores across most models indicate that the challenges are not merely due to basic formatting errors (\eg including an extra column or duplicate entries) but stem from fundamental graph matching issues. In addition, smaller models within the same family perform significantly worse (as seen in the \texttt{gpt-}, \texttt{llama-}, and \texttt{gemini-} series), highlighting the benchmark's effectiveness in differentiating LLM capabilities.

\begin{figure}[h]
    \centering
    \vspace{-0.5cm}\includegraphics[width=0.9\linewidth]{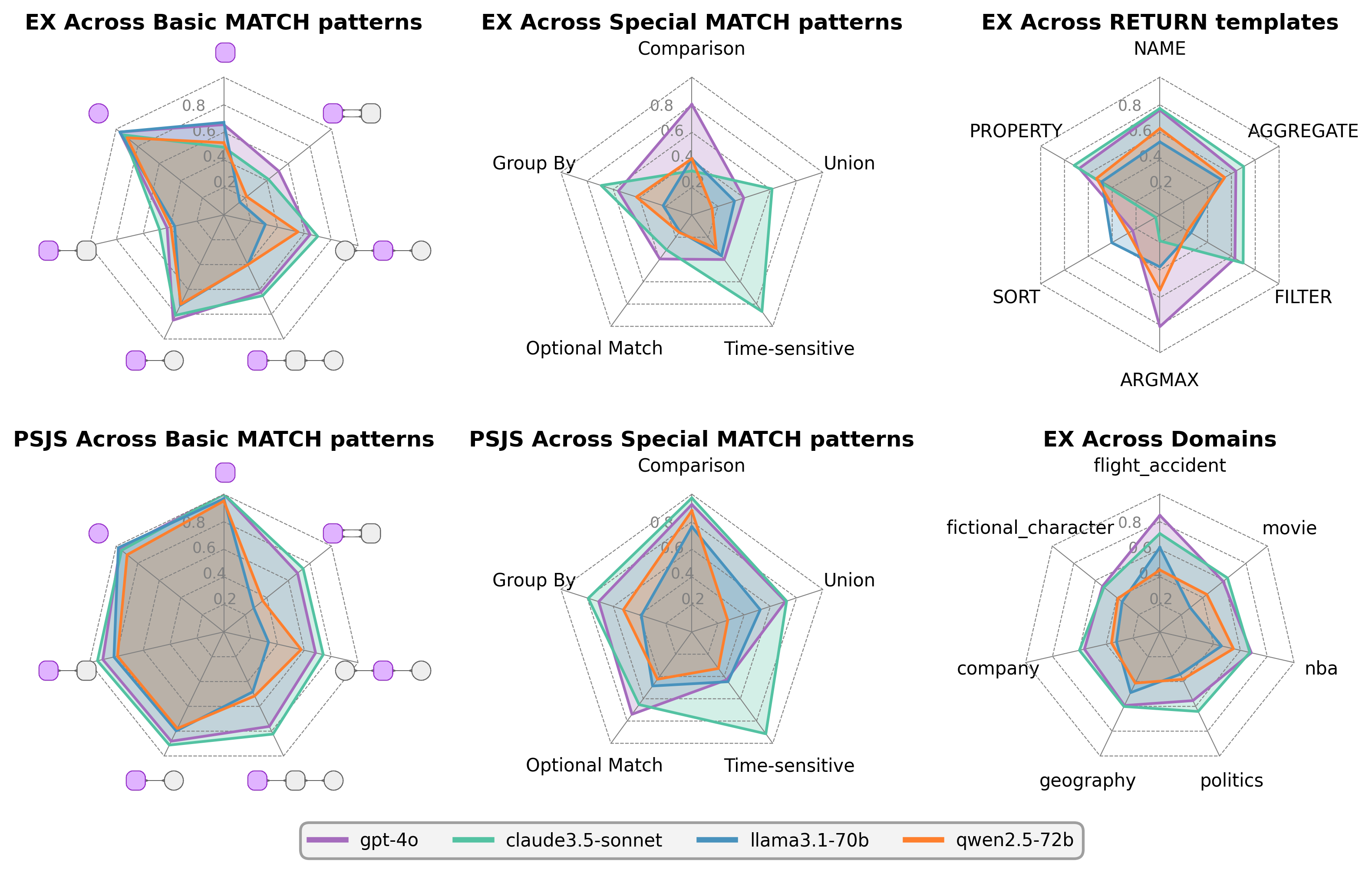}    \vspace{-0.1cm}\caption{Performance across basic and special \texttt{MATCH} patterns, \texttt{RETURN} templates and domains.}
    \label{fig:radar}\vspace{-0.1cm}
\end{figure}

\subsection{Performance Across Graph Matching Patterns}

Next, we analyze the performance breakdown across various dimensions. For this analysis, we focus on \texttt{gpt-4o}, \texttt{claude3.5-sonnet}, \texttt{qwen2.5-72b}, and \texttt{llama3.1-70b}, which represent the top 2 performing proprietary LLMs and the top 2 open-source LLMs.

In \autoref{fig:radar}, the four charts on the left show the execution accuracy and PSJS of these models across various graph matching patterns. Among the basic categories, all models exhibit similar trends—achieving near-perfect accuracy on pattern \raisebox{-0.1cm}{\includegraphics[height=0.4cm]{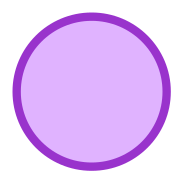}} while performing worst on pattern \raisebox{-0.1cm}{\includegraphics[height=0.4cm]{figures/match_patterns/basic_7.png}}. The PSJS chart, which evaluates graph matching alone, shows a consistent gradual decline in performance as the graph patterns include more relations.

Comparing the EX and PSJS charts provides insight into whether errors are caused by graph matching. For example, all models achieve near-perfect PSJS scores but low EX on pattern \raisebox{-0.1cm}{\includegraphics[height=0.4cm]{figures/match_patterns/basic_1.png}}. Upon manual inspection, we identified that most errors for this pattern result from incorrect deduplication—merging distinct entities that have the same name.

Within the special categories, models display varying weaknesses across different patterns. For instance, \texttt{gpt-4o} struggles with time-sensitive questions, whereas \texttt{claude3.5-sonnet} performs poorly on comparison questions.

\subsection{Performance Across \texttt{RETURN} Templates}

The top right chart in \autoref{fig:radar} displays execution accuracy across the \texttt{RETURN} templates. Here, the models also demonstrate different weaknesses depending on the template. Interestingly, \texttt{claude3.5-sonnet} achieves near-zero accuracy on \texttt{SORT} questions. Upon closer inspection, we observed that it frequently includes the variable used to rank the entities as an extra column, even though the questions only request the entity names, thus resulting in zero execution accuracy (however, PSJS is 1.0 in most of these cases since it is designed to be independent of the \texttt{RETURN} clause).

\subsection{Performance Across Domains}

The bottom right chart in \autoref{fig:radar} shows the execution accuracy across differnt domains. All four models exhibit similar trends, with \texttt{flight\_accident} and \texttt{nba} being the easiest, while showing comparable performance across the remaining domains.

\begin{figure}[h]
    \centering
    \vspace{-0.5cm}\includegraphics[width=0.8\linewidth]{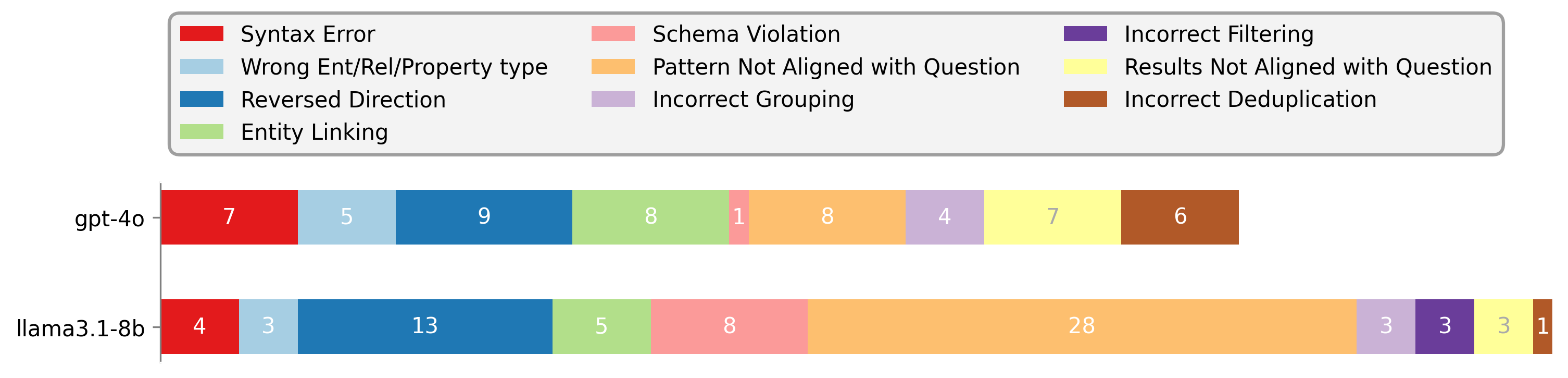}    \vspace{-0.2cm}\caption{Distribution of errors made by \texttt{gpt-4o} and \texttt{llama3.1-8b} on 50 randomly sampled incorrect predictions. Note that a model might make multiple errors on one instance.}
    % These numbers should not be interpreted as the absolute error counts for the entire test set when comparing across models.
    \label{fig:error_distribution}\vspace{-0.1cm}
\end{figure}

\subsection{Error Analysis}

We further conduct an error analysis to investigate the types of errors made by LLMs. Specifically, we focus on \texttt{gpt-4o} and \texttt{llama3.1-8b} to examine whether smaller models behave differently from larger ones. For each model, we randomly sample 50 task instances where they make incorrect predictions and annotate the errors observed in each instance. The error category taxonomy is developed during the annotation process (\aref{appendix:error_categories} shows the definitions of each error category with examples). Two sample predictions are shown in \autoref{tab:error_sample} and the distribution of error categories is shown in \autoref{fig:error_distribution}. Both models exhibit diverse errors spanning 10 distinct categories. Some error categories that frequently occur in both models include \texttt{Reversed Direction} (where the model reverses the direction of a relation), \texttt{Entity Linking} (where the entity name does not correspond to the intended entity in the database), and \texttt{Pattern Not Aligned with Question} (where the \texttt{MATCH} pattern conforms to the schema but does not align with the question's intent). Compared to \texttt{gpt-4o}, \texttt{llama3.1-8b} makes a significantly higher proportion of schema violation errors, indicating its inferior schema following capabilities.

\begin{table*}[h]
\centering
\scriptsize
\begin{tabular}{p{3.3cm}p{4.2cm}p{5.2cm}}
\toprule
\textbf{Question} & \textbf{Ground-truth Cypher} & \textbf{Predicted Cypher \& Errors} \\
\midrule
What are the names of lakes situated in countries through which the Natara River flows, with an area of less than 390,000 square kilometers? \hfill (\textit{geography}) & 
\vspace{-1.6ex}\begin{lstlisting}[style=TinyCypher]
MATCH (n:Lake)-[r0:locatedIn]->
(m0:Country)<-[r1:flowsThrough]-
(m1:River {name: 'Natara'}) 
WITH DISTINCT n 
WHERE n.area_km2 < 390000 
RETURN n.name
\end{lstlisting}\vspace{-1.6ex} & 
\vspace{-1.6ex}\begin{lstlisting}[style=TinyCypher]
MATCH (r:River {name: 'Natara River'})-
      [:flowsThrough]->(c:Country)<-
      [:locatedIn]-(l:Lake) 
WHERE l.area_km2 < 390000 
RETURN DISTINCT l.name
\end{lstlisting}\vspace{-1.6ex}
~

\textcolor{entitylinking}{$\blacksquare$ \texttt{Entity Linking:} } \texttt{\textquotesingle Natara River\textquotesingle} does not exist

\textcolor{incorrectdeduplication}{$\blacksquare$ \texttt{Incorrect Deduplication:}} Distinct rivers with the same name should not be merged

\hfill (\texttt{gpt-4o})
\\
\midrule
What are the names of film series that include movies originating from the Czech Republic? \hfill (\textit{movie}) & 
\vspace{-1.6ex}\begin{lstlisting}[style=TinyCypher]
MATCH (n:FilmSeries)<-
[r0:partOfSeries]-(m0:Movie)-
[r1:originatesFrom]->
(m1:Country {name: 'Czech Republic'}) 
WITH DISTINCT n 
RETURN n.name
\end{lstlisting}\vspace{-1.6ex} & 
\vspace{-1.6ex}\begin{lstlisting}[style=TinyCypher]
MATCH (m:Movie)-[:originatesFrom]->
      (c:Country {name: 'Czech Republic'})-
      [:partOfSeries]->(fs:FilmSeries) 
RETURN DISTINCT fs.name
\end{lstlisting}\vspace{-1.6ex}
~

\textcolor{schemaviolation}{$\blacksquare$ \texttt{Schema violation:}} \texttt{partOfSeries} between \texttt{Country} and \texttt{FilmSeries} is invalid  

\hfill (\texttt{llama3.1-8b})\\
\bottomrule
\end{tabular}
\caption{Sample predictions of \texttt{gpt-4o} and \texttt{llama3.1-8b} with annotated error categories.}
\label{tab:error_sample}
\end{table*}

% \yanlin{Can GPT-4 answer without graphs?}

% \yanlin{Show that our proposal is easier than direct QA over Wikidata and more efficient (in terms of latency)}

% \yanlin{Show provenance subgraph of random samples}

\section{Related Work} \label{sec:related}

\subsection{KBQA and Graph Retrieval Methods}

Our work is related to knowledge base question answering (KBQA) as CypherBench serves as a benchmark for evaluating KBQA and graph retrieval methods.

We categorize existing KBQA and graph retreival methods into two types (see \autoref{tab:graph_retrieval_methods}): approximate retrieval methods, which identify top relevant elements based on some notion of relevance to the question, and precise retrieval methods, which retrieve exactly what the question specifies by executing a formal language query. 

\begin{table*}[ht!]
\scriptsize
\centering
\begin{tabular}{p{5cm}p{8.0cm}}
\toprule
\textbf{Graph Retrieval Method} & \textbf{Papers / Open-source Projects} \\
\midrule
\rowcolor[gray]{0.9}
\multicolumn{2}{l}{\textit{Approximate Retrieval}} \vspace{2pt} \\
Entity linking + $k$-hop neighbourhood & \raisebox{-2pt}{\includegraphics[height=9pt]{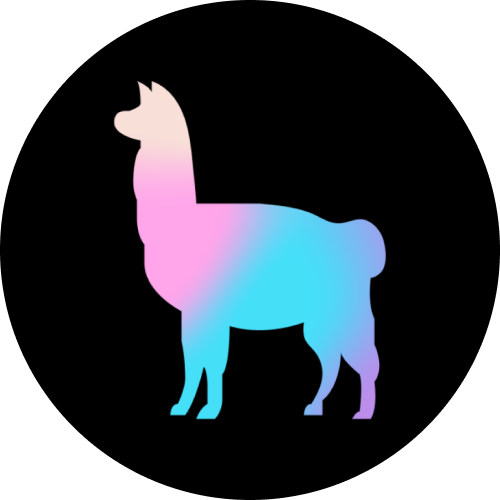}} LlamaIndex \cite{llamaindex2023propertygraph}, MCCNN \cite{dong-etal-2015-question}, UniK-QA \cite{oguz-etal-2022-unik}, CLOCQ \cite{christmann2022beyond}, Convinse \cite{christmann2022conversational}, Explaignn \cite{christmann2023explainable}, Temple-MQA \cite{cheng2024multi}, Subgraph Retriever \cite{zhang-etal-2022-subgraph}, QA-GNN \cite{yasunaga-etal-2021-qa}, MHGRN \cite{feng-etal-2020-scalable}, RoG \cite{luoreasoning}, UniKGQA \cite{jiangunikgqa}  \\
\addlinespace[0.5em]
Top-$k$ entities / paths / pseudo-docs & STaRK \cite{wu2024stark}, DiFaR \cite{baek-etal-2023-direct}, UDT-QA \cite{ma-etal-2022-open}, DecAF \cite{yu2022decaf}, PoG \cite{tan2024paths}, KARPA \cite{fang2024karpa} \\
\midrule
\rowcolor[gray]{0.9}
\multicolumn{2}{l}{\textit{Precise Retrieval}} \vspace{2pt} \\
Text-to-SPARQL through intermediate logical form (\eg $\lambda$-DCS, S-expression, etc.) & S-expression \cite{gu2021beyond}, $\lambda$-DCS \cite{berant-etal-2013-semantic, berant-liang-2014-semantic}, ComplexWebQ \cite{talmor-berant-2018-web}, Staged Query Graph \cite{yih-etal-2015-semantic}, Graph Query \cite{su-etal-2016-generating}, Abstract Query Graph \cite{chenformal}, KoPL \cite{cao-etal-2022-kqa}, GraphQ IR \cite{nie-etal-2022-graphq}, KB-BINDER \cite{li-etal-2023-shot} \\
\addlinespace[0.5em]
Text-to-SPARQL & \raisebox{-2pt}{\includegraphics[height=8pt]{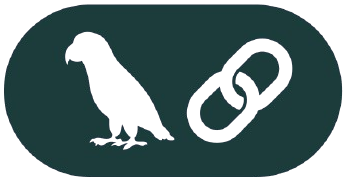}} Langchain \cite{chase2022langchain}, sparqlgen \cite{kovriguina2023sparqlgen}, T5-sparql \cite{banerjee2022modern}, sparql-llm \cite{emonet2024llm}, SPARKLE \cite{lee2024sparkle}, WikiSP \cite{xu-etal-2023-fine}, SPINACH \cite{liu-etal-2024-spinach} \\
\addlinespace[0.5em]
Text-to-Cypher (Our focus) & \raisebox{-2pt}{\includegraphics[height=9pt]{figures/logos/LlamaLogoBrowserTab.png}} LlamaIndex \cite{llamaindex2023propertygraph}, \raisebox{-2pt}{\includegraphics[height=8pt]{figures/logos/langchain-removebg-preview.png}} Langchain \cite{chase2022langchain}, UniOQA \cite{li2024unioqa}   \\
\bottomrule
\end{tabular}
\caption{Graph retrieval methods adopted by existing research papers and open-source projects.}
\label{tab:graph_retrieval_methods}
\end{table*}

The most common approximate retrieval method involves retrieving the $k$-hop neighborhood of the entities mentioned in the question \cite{dong-etal-2015-question, oguz-etal-2022-unik, christmann2022beyond, christmann2022conversational, christmann2023explainable, cheng2024multi, zhang-etal-2022-subgraph, yasunaga-etal-2021-qa, feng-etal-2020-scalable, luoreasoning, jiangunikgqa}. Another line of work verbalizes entities or relations into text and uses embedding-based methods to retrieve the top-$k$ most relevant elements \cite{wu2024stark,baek-etal-2023-direct,ma-etal-2022-open,yu2022decaf}. The fundamental limitation of these methods is their inability to handle questions involving a large number of entities (\ie global queries or complex aggregation queries)\sajhidden{we need to define these query types somewhere.}, as they require processing all the retrieved information during the answer generation step. Additionally, these methods usually rely on expensive in-memory operations or require embedding the entire graph, which limits their feasibility when applied to full-scale modern knowledge graphs which typically contain billions of triples.

Precise retrieval methods translate the question into a formal language query that fetches exactly what the question asks for. However, most approaches in this category are based on custom logical forms, which are either transpiled into actual database queries or executed by a custom engine \cite{gu2021beyond, berant-etal-2013-semantic, berant-liang-2014-semantic, talmor-berant-2018-web, yih-etal-2015-semantic, su-etal-2016-generating, chenformal, cao-etal-2022-kqa, nie-etal-2022-graphq, li-etal-2023-shot}. These custom logical forms are easier to generate for pre-LLM models due to their simpler syntax, but often lack support for certain graph querying features like grouping and variable-length path matching. The ones that are executed by custom engines also face limitations in scalability and real-world applicability compared to standard database query languages. For example, the recently proposed KoPL \cite{cao-etal-2022-kqa} queries are executed by loading and processing the entire graph in-memory, which makes it impractical to handle graphs of a size comparable to Wikidata. While some recent works use LLMs to directly generate graph database queries (\eg SPARQL or Cypher) \cite{kovriguina2023sparqlgen, banerjee2022modern, emonet2024llm, lee2024sparkle, li2024unioqa, xu-etal-2023-fine, liu-etal-2024-spinach}, they often make simplifications such as assuming that identifiers are provided or working with smaller graphs. Notably, the recently introduced SPINACH \cite{liu-etal-2024-spinach} operates over full Wikidata using an agentic workflow.

% \sajhidden{where does MetaQA-cypher fit here? or KQAPro? we need to also discuss differences with existing benchmarks.}

\subsection{Text-to-Query and KBQA Benchmarks}

CypherBench takes the form of a text-to-query benchmark, consisting of databases along with (question, database query) pairs. KBQA benchmarks represent a specific type of text-to-query benchmarks, where the databases are knowledge graphs, and the queries are graph database queries. In \autoref{tab:nl2q_overview}, we compare CypherBench with current representative text-to-query benchmarks.

Looking at the Schema Size and Data Size columns provides insights into the complexity of the databases in existing text-to-query benchmarks, both in terms of their schemas and stored data. Most existing KBQA benchmarks \cite{berant-etal-2013-semantic,moon-etal-2019-opendialkg,dubey2019lc,gu2021beyond,cao-etal-2022-kqa} are predominantly based on text-to-SPARQL over RDF knowledge graphs. However, as discussed in \autoref{sec:rdf-to-property}, the massive schema of RDF knowledge graphs poses significant challenges when using these benchmarks to evaluate LLMs in zero-shot settings. In contrast, the graphs in CypherBench have a schema size comparable to those in text-to-SQL benchmarks \cite{yu-etal-2018-spider,li2024can}, while still encompassing up to 7 million entities.

\begin{table*}[ht]
\scriptsize
\centering
\scalebox{0.94}{\begin{tabular}{llrllc}
\toprule
\textbf{Benchmark} & \textbf{Data Source} & \textbf{\#graph/db} & \textbf{Avg. Schema Size} {\tiny (per graph/db)} & \textbf{Data Size} & \textbf{LLM Efficient?} \\
\midrule
\rowcolor[gray]{0.9}
\multicolumn{6}{l}{\textit{Text-to-SQL / relational data}} \vspace{2pt} \\
Spider \cite{yu-etal-2018-spider} & Wikipedia, etc. & 200 & 5.1 tables, 27.6 columns & 400k rows & \checkmark \\
BIRD-SQL \cite{li2024can} & Kaggle, etc. & 95 & 7.3 tables, 54.2 columns & 52M rows & \checkmark \\
\midrule
\rowcolor[gray]{0.9}
\multicolumn{6}{l}{\textit{Text-to-SPARQL / RDF graphs}}\vspace{2pt} \\
LC-Quad 2.0 \cite{dubey2019lc}   & Wikidata & 1 & 12k relation types & 114M entities & \texttimes \\
GrailQA \cite{gu2021beyond}   & Freebase & 1 & 37k relation types & 45M entities & \texttimes \\
KQA Pro \cite{cao-etal-2022-kqa} & FB15k-237 & 1 & 0.8k relation types & 16k entities & \texttimes \\
\midrule
\rowcolor[gray]{0.9}
\multicolumn{6}{l}{\textit{Text-to-nGQL / property graphs}}\vspace{2pt} \\
$R^3$-NL2GQL$^\dagger$ \cite{zhou2024r} & OpenKG & 3 & 5.3 relation types, 13 properties & 46k entities & \checkmark \\
Fin/Medi-GQL$^\dagger$ \cite{liang2024aligning} & OpenKG & 2 & 13 relation types, 38 properties & 713k entities & \checkmark \\
\midrule
\rowcolor[gray]{0.9}
\multicolumn{6}{l}{\textit{Text-to-Cypher / property graphs}}\vspace{2pt} \\
MetaQA-Cypher \cite{nie-etal-2022-graphq} & OMDb & 1 & 5 relation types, 5 properties & 43k entities & \checkmark \\
SpCQL$^\dagger$ \cite{guo2022spcql} & OwnThink & 1 & 480k relation types, 1 property & 16M entities & \texttimes \\
Neo4j Text2Cypher (2024) & neo4j-graph-examples & - & - & - & \checkmark \\
CypherBench (ours)  & Wikidata & 11 & 7.5 relation types, 18.7 properties & 7.8M entities & \checkmark \\
\bottomrule
\end{tabular}}
\caption{Comparison of representative text-to-query benchmarks. Benchmarks marked by $\dagger$ are non-English ($R^3$-NL2GQL, FinGQL, MediGQL and spQCL are in Chinese). The column ``LLM Efficient?'' refers to whether the database schema from the benchmark can fit in the typical context window of LLMs. Existing KBQA benchmarks pose challenges for evaluation in zero-shot settings with LLMs due to the massive schema of RDF knowledge graphs.}
\label{tab:nl2q_overview}
\end{table*}

\begin{table*}[!htb]
\scriptsize 
\centering
\scalebox{0.92}{\begin{tabular}{p{1.7cm} cccc}
\toprule
 & Natural Questions & HotpotQA & KQA Pro & MetaQA-Cypher \\
  & \textit{(single-hop Text QA)} & {\textit{(multi-hop Text QA)}} & \textit{(text-to-SPARQL)} & \textit{(text-to-Cypher)} \\
\midrule
\rowcolor[gray]{0.9}
\multicolumn{5}{l}{\textit{CypherBench Basic Graph Patterns}} \vspace{2pt} \\
\raisebox{-0.3\height}{\includegraphics[height=0.4cm]{figures/match_patterns/basic_1.png}} & &  & 
& \\
\raisebox{-0.3\height}{\includegraphics[height=0.4cm]{figures/match_patterns/basic_2.png}} & \checkmark &  & \checkmark
& \checkmark \\
\raisebox{-0.3\height}{\includegraphics[height=0.4cm]{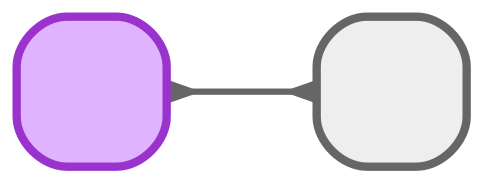}} & &  & 
& \\
\raisebox{-0.3\height}{\includegraphics[height=0.4cm]{figures/match_patterns/basic_4.png}} & \checkmark & \checkmark & \checkmark
& \checkmark \\
\raisebox{-0.3\height}{\includegraphics[height=0.4cm]{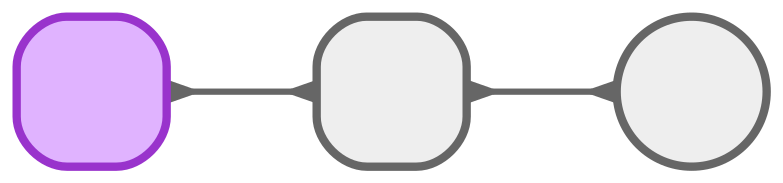}} & & \checkmark & \checkmark
& \\
\raisebox{-0.3\height}{\includegraphics[height=0.4cm]{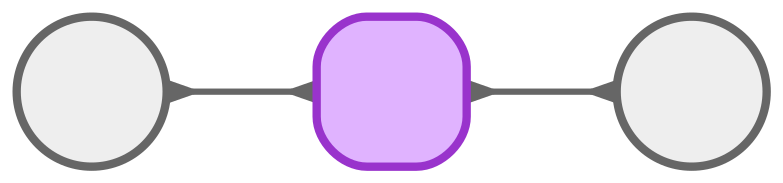}} & &  & 
& \\
\raisebox{-0.3\height}{\includegraphics[height=0.4cm]{figures/match_patterns/basic_7.png}} & &  & 
& \\
\midrule
\rowcolor[gray]{0.9}
\multicolumn{5}{l}{\textit{CypherBench Special Graph Patterns}} \vspace{2pt} \\
\raisebox{-0.3\height}{\includegraphics[height=0.4cm]{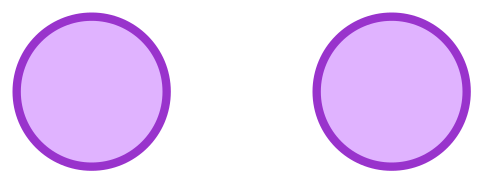}} & & \checkmark & \checkmark
& \\
\raisebox{-0.3\height}{\includegraphics[height=0.4cm]{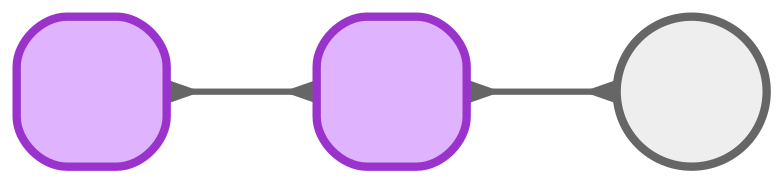}} & &  & 
& \\
\raisebox{-0.3\height}{\includegraphics[height=0.4cm]{figures/match_patterns/optional.png}} & &  & 
& \\
\raisebox{-0.3\height}{\includegraphics[height=0.4cm]{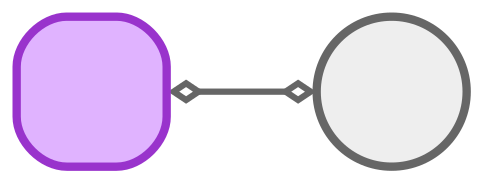}} & &  & \checkmark
& \\
\raisebox{-0.3\height}{\includegraphics[height=0.4cm]{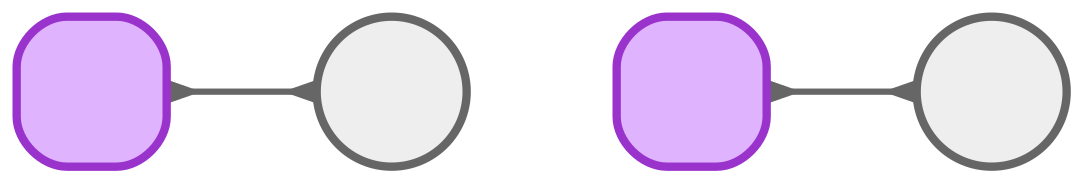}} & &  & 
& \\
\bottomrule
\end{tabular}}
\caption{ Graph matching patterns covered by previous question answering datasets. We also include Natural Questions \cite{kwiatkowski-etal-2019-natural} and HotpotQA \cite{yang-etal-2018-hotpotqa} as representative knowledge-intensive single-hop and multi-hop text QA datasets here. We determined the patterns based on the question curation approach described in the paper and 100 randomly sampled questions. }
\label{tab:benchmark_compare}
\end{table*}

In recent years, several benchmarks focusing on text-to-Cypher or text-to-nGQL\footnote{nGQL is the query language for NebulaGraph, a property graph database.} have been proposed \cite{nie-etal-2022-graphq,guo2022spcql,zhou2024r,liang2024aligning,zhong2024synthet2c}. MetaQA-Cypher \cite{nie-etal-2022-graphq} and SpCQL \cite{guo2022spcql} are the earliest efforts to develop text-to-Cypher benchmarks. MetaQA-Cypher is adapted from MetaQA \cite{zhang2018variational}, a KBQA dataset built on a movie knowledge graph, with Cypher queries annotated using rule-based methods. SpCQL is based on OwnThink\footnote{OwnThink is provided in a plain triple format, lacking the notion of entity types as well as entity and relation properties. When stored in Neo4j, it uses a single entity type, \texttt{ENTITY}, and a single relation type, \texttt{Relationship}, with the actual relation type stored in the \texttt{name} property of the relations. Consequently, OwnThink resembles RDF graphs more closely than property graphs.}, a Chinese encyclopedic knowledge graph. The questions in SpCQL were collected from online forums and annotated with Cypher queries by database professionals. \cite{zhou2024r} constructed a text-to-nGQL dataset over three domain knowledge graphs through a combination of human curation and LLM generation. \cite{liang2024aligning} instead employed a templated generation approach using eight human-curated templates.  However, these benchmarks are restricted to a small number of domains (with the exception of SpCQL), and their questions lack diversity, covering only a limited number of the graph matching patterns in CypherBench (as shown in \autoref{tab:benchmark_compare}).

A parallel effort to create a large-scale text-to-Cypher benchmark is the Neo4j Text2Cypher (2024) dataset\footnote{\url{https://huggingface.co/datasets/neo4j/text2cypher-2024v1}}\cite{ozsoy2024text2cypher}. This dataset was developed by cleaning and combining 25 public datasets from Neo4j internal projects, Hugging Face, and academic papers. Compared to previous benchmarks, it is significantly more diverse in terms of domains and question types. However, 49\% of the questions in the dataset are not linked to any actual graphs (many of these questions are synthetically generated and do not have a corresponding database). Instead, the dataset only provides a textual description of the graph schema for text-to-Cypher generation. This makes it impossible to execute the Cypher queries to evaluate execution-based metrics like EX and PSJS. The remaining 51\% of the questions is based on demo graphs from the neo4j-graph-examples repositories\footnote{\url{https://github.com/neo4j-graph-examples}}. These demo graphs are typically smaller in size and do not comprehensively cover all domain entities, unlike CypherBench, which provides full coverage of domain entities.

\subsection{GraphRAG} 
% \footnote{It is worth noting that the original GraphRAG system in \cite{edge2024local} uses a graph structure slightly different from a typical knowledge graph, where the nodes include entities and communities at multiple levels, with retrieval performed by fetching all communities at a specific level.}

% Recently, Microsoft introduced GraphRAG \cite{edge2024local} to address corpus-level summarization queries (\eg ``What are the main themes in the dataset?'', similar to the global queries we explored in this work) that cannot be handled by standard top-k embedding-based retrieval. At a high-level, GraphRAG utilizes a knowledge graph to index textual documents to handle query that depends on a large amount of documents.

Recently, Microsoft introduced GraphRAG \cite{edge2024local} to address corpus-level summarization queries (\eg “What are the main themes in the dataset?”), which are similar to the global queries explored in this work and cannot be handled by standard top-k embedding-based retrieval methods. At a high level, GraphRAG leverages a centralized knowledge graph to index textual documents, enabling it to handle queries that rely on a large volume of documents. The GraphRAG system has two main stages: knowledge graph construction during indexing time and graph retrieval during query time \cite{peng2024graph}. It is worth noting that the original GraphRAG system in \cite{edge2024local} uses a graph formalism slightly different from a typical knowledge graph, where the nodes are entity communities at various abstraction levels, with retrieval performed by fetching all communities at a specific level. 

Subsequently, LlamaIndex, the leading open-source LLM framework for RAG workflows, introduced the Property Graph Index \cite{llamaindex2023propertygraph} for general-purpose question answering. It constructs a Neo4j property graph from textual documents using LLMs during indexing time, and conducts graph retrieval via text-to-Cypher during query time. Our work provides the first comprehensive text-to-Cypher benchmark for evaluating graph retrieval, a critical component in GraphRAG.

\subsection{Mapping RDF to Property Graphs}

Several studies from the semantic web community have explored methods for transforming RDF graphs into property graphs \cite{hartig2014reconciliation,tomaszuk2016rdf,schatzle2016s2x,angles2020mapping,matsumoto2018mapping}\footnote{For a more detailed survey, we refer readers to \cite{angles2019rdf}.}. However, many of these methods require processing the entire RDF dump, which can be computationally expensive for full-size modern RDF graphs like Wikidata. For example, \cite{hartig2014reconciliation} proposed a two-step process for RDF*, an RDF extension: first, RDF triples are mapped directly to edges in the property graph, and then edges that represent entity properties are transformed into node properties. An exception is \cite{matsumoto2018mapping} \footnote{\url{https://github.com/g2glab/g2g}}, which adopts an approach similar to ours by transforming RDF into property graphs through executing SPARQL queries over RDF. However, their method lacks key functionalities described in \autoref{sec:graph} which are essential for ensuring the output quality.

\subsection{Knowledge Graph Subsetting}

There are also a few tools developed to extract domain-specific subgraphs of Wikidata or general RDF knowledge graphs to tackle the scalability challenges of modern knowledge graphs\footnote{Note that this differs from subgraph retrieval in approximate graph retrieval methods, where a smaller subgraph is extracted for each question.}. For example, KGTK \cite{ilievski2020kgtk}, WDumper\footnote{\url{https://github.com/bennofs/wdumper}}, and WDSub\footnote{\url{https://github.com/weso/wdsub}} are a few such tools \cite{hosseini2023wikidata}. However, these tools also operate by processing the entire RDF dump and produce RDF as output.

% \yanlin{Querying Wikidata: Comparing SPARQL,
% Relational and Graph Databases}
\section{Conclusion}

% The goals of this work are two-fold: 1) the development of a novel methodology to enable efficient and accurate text-to-Cypher retrieval over modern RDF knowledge graphs such as Wikidata, and 2) the creation of a large-scale text-to-Cypher / KBQA dataset that includes high-quality property graphs and diverse questions. 

Since its inception, Wikidata has received over 2 billion edits by users worldwide and continues to be actively maintained by over 42,000 editors in the past year, making it one of the most comprehensive knowledge sources available today. This study offers a viable pathway for integrating full-scale modern knowledge graphs like Wikidata with LLMs. The techniques we proposed, along with the numerous design choices made throughout this study, are all centered around accomplishing this goal. We believe our work offers new research opportunities in the areas of knowledge graphs and large-scale graph retrieval.

% \newpage
\clearpage

\bibliographystyle{unsrt}
\bibliography{bib/anthology2023, bib/custom}
% \newpage
\clearpage

\appendix

\section{Additional Technical Details}

\subsection{Graph Matching Patterns and \texttt{RETURN} Templates} 

The complete list of graph matching patterns is shown in \autoref{tab:graph_patterns} and the complete list of \texttt{RETURN} templates is shown in \autoref{tab:return_patterns}. 

\subsection{Question Rewriting Prompt} 

The prompt used to rewrite template-generated questions into more natural-sounding ones is presented in \autoref{tab:rewrite_prompt}.

\subsection{Text-to-Cypher Prompt} \label{appendix:text2cypher_prompt}

The text-to-Cypher prompt for evaluating LLMs is shown in \autoref{tab:prompt}.

\subsection{Schema Fetching in Neo4j} \label{appendix:schema}

In a practical text-to-Cypher scenario where only the Neo4j database endpoint (host and port) is provided, the graph schema must be retrieved from the database by executing certain Cypher queries. Neo4j provides built-in procedures such as \texttt{db.schema.visualization} and \texttt{apoc.meta.data} for this purpose. However, we observed that both methods yield inaccurate results when applied to large graphs: \texttt{db.schema.visualization} may return non-existent relationships, while \texttt{apoc.meta.data} can miss certain relationships. To address this issue, we use the following queries to retrieve the schemas: \newline 
\begin{minipage}[t]{0.47\textwidth}
    \vspace{-5pt}
        \begin{tcolorbox}[title={\scriptsize Cypher for fetching entity property schemas}, boxrule=0.8pt, colframe=gray, boxsep=2pt, left=2pt, right=2pt, top=2pt, bottom=2pt]
            \begin{lstlisting}[style=TinyCypher]
MATCH (n)
UNWIND labels(n) AS label
WITH label, keys(n) AS propertyKeys, n
UNWIND propertyKeys AS property
WITH DISTINCT label, property, apoc.meta.cypher.type(n[property]) as type
WITH label AS label, apoc.coll.sortMaps(collect({
      property:property, type:type}), 'property') AS properties
RETURN label, properties ORDER BY label
\end{lstlisting}
        \end{tcolorbox}
    \end{minipage}\hspace{0.01\textwidth}
    \begin{minipage}[t]{0.47\textwidth}
    \vspace{-5pt}
        \begin{tcolorbox}[title={\scriptsize Cypher for fetching relation property schemas}, boxrule=0.8pt, colframe=gray, boxsep=2pt, left=2pt, right=2pt, top=2pt, bottom=2pt]
            \begin{lstlisting}[style=TinyCypher]
MATCH ()-[r]-()
WITH type(r) AS type, keys(r) AS propertyKeys, r
UNWIND propertyKeys AS property
WITH DISTINCT type, property, apoc.meta.cypher.type(r[property]) AS propType
WITH type, apoc.coll.sortMaps(collect({
      property:property, type:propType}), 'property') AS properties
RETURN type, properties ORDER BY type
\end{lstlisting}
        \end{tcolorbox}
    \end{minipage} 

\vspace{0.15cm}
    
    \begin{minipage}[t]{0.47\textwidth}
    \vspace{-5pt}
        \begin{tcolorbox}[title={\scriptsize Cypher for fetching relation schemas}, boxrule=0.8pt, colframe=gray, boxsep=2pt, left=2pt, right=2pt, top=2pt, bottom=2pt]
            \begin{lstlisting}[style=TinyCypher]
MATCH (n)-[r]->(m)
UNWIND labels(n) AS start UNWIND labels(m) AS end
RETURN DISTINCT start, type(r) as type, end
ORDER BY type, start, end
\end{lstlisting}
        \end{tcolorbox}
    \end{minipage} 

\vspace{0.07cm}
The results are then aggregated and serialized into the format shown in \autoref{tab:prompt}. While these queries are less efficient than the built-in procedures (approximately 15 times slower than \texttt{apoc.meta.data}), they produce complete and accurate schemas deterministically.

\subsection{Text-to-Cypher Error Taxonomy} \label{appendix:error_categories}

\autoref{tab:error_definition} shows the detailed definitions of each error category with examples.

\section{Additional CypherBench Statistics} \label{appendix:cypherbench}

The graph statistics are shown in \autoref{tab:graph_stats}. The schemas of the 11 property graphs are shown in \autoref{tab:all_graphs} and \autoref{tab:all_graphs_1}. 

\begin{table*}[ht]
\scriptsize 
\centering
\begin{tabular}{lrrrrrr}
\toprule
{Graph} & {Ent.} & {Rel.} & {Ent. Types} & {Rel. Types} & {Properties} & {Wikipedia} \\
\midrule
\textit{art}                 & 1.1M   & 1.3M   & 6  & 8  & 16  & 64.9k  \\
\textit{biology}             & 3.7M   & 7.5M   & 4  & 5  & 8   & 447.7k \\
\textit{company}             & 581.3k & 299.6k & 4  & 6  & 14  & 166.3k \\
\textit{fictional character} & 28.9k  & 40.5k  & 4  & 11 & 12  & 8.5k   \\
\textit{flight accident}     & 1.7k   & 2.2k   & 5  & 5  & 25  & 1.6k   \\
\textit{geography}           & 773.5k & 903.8k & 8  & 12 & 19  & 73.1k  \\
\textit{movie}               & 459.4k & 1.9M   & 7  & 9  & 21  & 262.2k \\
\textit{nba}                 & 4.3k   & 19.0k  & 7  & 7  & 27  & 4.3k   \\
\textit{politics}            & 885.2k & 1.5M   & 6  & 11 & 25  & 414.2k \\
\textit{soccer}              & 275.2k & 1.1M   & 6  & 5  & 26  & 206.6k \\
\textit{terrorist attack}    & 1.6k   & 1.5k   & 5  & 4  & 13  & 1.3k   \\
\midrule
{Total}               & 7.8M   & 14.7M  & 62 & 83 & 206 & 1.7M    \\
\bottomrule
\end{tabular}
\caption{ Statistics of the graphs. \textit{Wikipedia} refers to the number of English Wikipedia articles linked from the entities (roughly the number of entities with Wikipedia articles). }
\label{tab:graph_stats}
\end{table*}

\begin{table*}[h]
\vspace{-5pt}
\centering
\begin{tcolorbox}[title={\scriptsize Question Rewriting Prompt}, boxrule=0.8pt, colframe=gray, boxsep=2pt, left=2pt, right=2pt, top=2pt, bottom=2pt]
            \begin{lstlisting}[style=TinyJSON]
Rewrite the given template-generated question in a text-to-Cypher translation task to make it sound more natural:
- Ensure the rewritten question remains semantically equivalent to the original question and the provided Cypher query. Do not remove or add any constraints.
  - Pay attention to the direction of the relation pattern (indicated by `->` or `<-`) in the Cypher query.
    For example, `(n:Character)-[r0:hasFather]->(m0:Character)` indicates m0 is the father of n,
    while `(n:Character)<-[r0:hasFather]-(m0:Character)` matches n as the father of m0.
  - Pay attention to the direction of relation in the template-generated question.
    For example `List the names of Character that "Rhaenys Targaryen" hasFather"` means "Rhaenys Targaryen" connects 
    to the Character via relation `hasFather`, thus the questions is asking for the father of Rhaenys Targaryen.
    While `List the names of Character that hasFather "Rhaenys Targaryen"` selects the Character that has Rhaenys Targaryen as father.
- Ensure the rewritten question is grammatically correct and sounds natural.
- The brackets in the question are parsing hints for the question structure to ensure it is unambiguous. Do not include them in the rewritten question.
- For relation types (e.g. hasCastMember), rewrite them to natural language and diversify the expressions. Feel free to change from passive to active voice or vice versa.
  - e.g. A hasCastMember B -> B is cast in A, B stars in A, A features B, etc.
- For entity types (e.g. Person, FlightAccident) and properties (e.g. watershed_area_km2) rewrite them to natural language and diversify the expressions.
  - e.g. Person -> individual; human; passenger; etc. 
  - e.g. TaxonRank -> taxonomic rank; etc.
  - e.g. FlightAccident -> aviation accident; plane crash; etc.
  - e.g. watershed_area_km2 -> size of the watershed in square kilometers; area covered by the watershed in km^2; etc.
- For multi-hop patterns, you can simplify it if the same meaning is preserved.
  - e.g. "teams that belong to a division that belongs to Western Conference" -> "teams in the Western Conference"
  - e.g. "the father of the mother of the person" -> "the person's maternal grandfather"
  - e.g. "the children of the father of the person" -> "the person's paternal siblings"
- For quoted names and string values, remove the quotes but ensure the same text is preserved.
- For numerical values and dates, diversify the expressions, but ensure the same value is preserved.
  - e.g. 1990-07-04 -> July 4th, 1990; 4 July 1990; 07/04/1990 (US format); 4th of July, 1990; etc.
  - e.g. 2000 -> two thousand; 2000; 2,000; 2k; etc.
- For operators (e.g. >, >=, IN, NOT IN, etc.), rewrite them to natural language and diversify the expressions but ensure the meaning is preserved.
  - e.g. NOT 'France' IN n.country_of_citizenship -> "is not a citizen of France"
- For "ascending" and "descending", rewrite them to natural language and diversify the expressions.
  - e.g. "the years in descending order" -> "the years from the most recent to the oldest"
- Output only the rewritten question, without any additional explanation.

=== Example ===
Cypher: MATCH (n:Taxon)<-[r0:hasParent]-(m0:Taxon)-[r1:hasConservationStatus]->(m1:ConservationStatus {name: 'Near Threatened'}) WITH DISTINCT n RETURN n.name
question: List the names of Taxon that [some Taxon that hasConservationStatus "Near Threatened"] hasParent
rewritten_question: What are the names of parents of taxa with a conservation status of Near Threatened?

Cypher: MATCH (n:Character)<-[r0:hasFather]-(m0:Character),(n:Character)<-[r1:killedBy]-(m0:Character) WITH DISTINCT n RETURN n.name
question: List the names of Character that a Character [hasFather and killedBy].
rewritten_question: List the names of fathers who killed their children.

=== Your task ===
Cypher: MATCH (n:Continent) WITH DISTINCT n RETURN n.name
question: List the names of Continent
rewritten_question:
\end{lstlisting}
\end{tcolorbox}
\caption{A sample question rewriting prompt.} \label{tab:rewrite_prompt}
\end{table*}

\begin{table*}[h]
\vspace{-5pt}
\centering
\begin{tcolorbox}[title={\scriptsize Text-to-Cypher Prompt}, boxrule=0.8pt, colframe=gray, boxsep=2pt, left=2pt, right=2pt, top=2pt, bottom=2pt]
            \begin{lstlisting}[style=TinyJSON]
Translate the question to Cypher query based on the schema of a Neo4j knowledge graph.
- Output the Cypher query in a single line, without any additional output or explanation. Do not wrap the query with any formatting like ```.
- Perform graph pattern matching in the `MATCH` clause if possible.
- Avoid listing the same entity multiple times in the results. However, if multiple distinct entities share the same name, their names should be repeated as separate entries.
- Do not return node objects. Instead, return entity names or properties.

Graph Schema:
{
  "name": "company",
  "entities": [
    {
      "label": "Company",
      "properties": {
        "launch_year": "int", "name": "str"
      }
    },
    {
      "label": "Country",
      "properties": {
        "name": "str"
      }
    },
    {
      "label": "Industry",
      "properties": {
        "name": "str"
      }
    },
    {
      "label": "Person",
      "properties": {
        "country_of_citizenship": "list[str]", "date_of_birth": "date", "date_of_death": "date", "gender": "str", "name": "str", "place_of_birth": "str"
      }
    }
  ],
  "relations": [
    {
      "label": "basedIn",
      "subj_label": "Company",
      "obj_label": "Country",
      "properties": {}
    },
    {
      "label": "foundedBy",
      "subj_label": "Company",
      "obj_label": "Person",
      "properties": {}
    },
    {
      "label": "hasBoardMember",
      "subj_label": "Company",
      "obj_label": "Person",
      "properties": {
        "end_year": "int", "start_year": "int"
      }
    },
    {
      "label": "hasCEO",
      "subj_label": "Company",
      "obj_label": "Person",
      "properties": {
        "end_year": "int", "start_year": "int"
      }
    },
    {
      "label": "operatesIn",
      "subj_label": "Company",
      "obj_label": "Industry",
      "properties": {}
    },
    {
      "label": "subsidiaryOf",
      "subj_label": "Company",
      "obj_label": "Company",
      "properties": {}
    }
  ]
}

Question: Provide the names of individuals who have served as board members for companies based in Russia, along with the count of such companies for each person.
Cypher: 
\end{lstlisting}
\end{tcolorbox}
\caption{A sample text-to-Cypher prompt used in experiments.} \label{tab:prompt}
\end{table*}

\begin{table*}[h]
\scriptsize 
\vspace{-5pt}
\begin{adjustwidth}{-2in}{-2in}
\centering
\begin{tabular}{p{2.5cm} p{5.5cm} p{7.5cm}}
\toprule
\textbf{Graph Pattern} & \textbf{Sample Question} & \textbf{Cypher Query} \\
\midrule
\textit{Basic Categories} \\
\midrule
\raisebox{-0.6\height}{\includegraphics[height=0.4cm]{figures/match_patterns/basic_1.png}} & $Q1.$ What are the names of terrorist attacks that occurred before March 13th, 1997? \hfill (\textit{terrorist~attack})  & \vspace{-1.6ex}\begin{lstlisting}[style=TinyCypher]
MATCH (n:TerroristAttack) WITH DISTINCT n
WHERE n.date < date('1997-03-13') RETURN n.name
\end{lstlisting}\vspace{-1.6ex}
\\
\midrule
\raisebox{-0.6\height}{\includegraphics[height=0.4cm]{figures/match_patterns/basic_2.png}} & $Q2.$ What is the discharge rate in cubic meters per second of the Guamués River? \hfill (\textit{geography}) & 
\vspace{-1.6ex}\begin{lstlisting}[style=TinyCypher]
MATCH (n:River {name: 'Guamués River'}) WITH DISTINCT n
RETURN n.discharge_m3_s
\end{lstlisting}\vspace{-1.6ex}
\\
\midrule
\raisebox{-0.6\height}{\includegraphics[height=0.4cm]{figures/match_patterns/basic_3.png}} & $Q3.$ List the players who have received an award, from tallest to shortest. \hfill (\textit{soccer}) & 
\vspace{-1.6ex}\begin{lstlisting}[style=TinyCypher]
MATCH (n:Player)-[r0:receivesAward]->(m0:Award)
WITH DISTINCT n RETURN n.name
ORDER BY n.height_cm DESC
\end{lstlisting}\vspace{-1.6ex}
\\
\midrule
\raisebox{-0.6\height}{\includegraphics[height=0.4cm]{figures/match_patterns/basic_4.png}} & $Q4.$ What are the names of taxa that feed on Synsphyronus lathrius? \hfill (\textit{biology}) & 
\vspace{-1.6ex}\begin{lstlisting}[style=TinyCypher]
MATCH (n:Taxon)-[r0:feedsOn]->(m0:Taxon {name: 'Synsphyronus lathrius'})
WITH DISTINCT n RETURN n.name
\end{lstlisting}\vspace{-1.6ex}
\\
\midrule
\raisebox{-0.6\height}{\includegraphics[height=0.4cm]{figures/match_patterns/basic_5.png}} & $Q5.$ What are the names of companies that operate in the same industries as Bardel Entertainment? \hfill (\textit{company}) & 
\vspace{-1.6ex}\begin{lstlisting}[style=TinyCypher]
MATCH (n:Company)-[r0:operatesIn]->(m0:Industry)<-
      [r1:operatesIn]-(m1:Company {name: 'Bardel Entertainment'})
WITH DISTINCT n RETURN n.name
\end{lstlisting}\vspace{-1.6ex}
\\
\midrule
\raisebox{-0.6\height}{\includegraphics[height=0.4cm]{figures/match_patterns/basic_6.png}} & $Q6.$ Who are the point guards who have played for the Toronto Raptors? \hfill (\textit{nba}) & 
\vspace{-1.6ex}\begin{lstlisting}[style=TinyCypher]
MATCH (n:Player)-[r0:playsFor]->(m0:Team {name: 'Toronto Raptors'}),(n:Player)-
      [r1:playsPosition]->(m1:Position {name: 'point guard'})
WITH DISTINCT n RETURN n.name
\end{lstlisting}\vspace{-1.6ex}
\\
\midrule
\raisebox{-0.6\height}{\includegraphics[height=0.4cm]{figures/match_patterns/basic_7.png}} & $Q7.$ What are the unique countries of citizenship of individuals who both wrote and acted in the same movie? \hfill (\textit{movie}) & 
\vspace{-1.6ex}\begin{lstlisting}[style=TinyCypher]
MATCH (n:Person)<-[r0:writtenBy]-(m0:Movie),
      (n:Person)<-[r1:hasCastMember]-(m0:Movie)
WITH DISTINCT n
UNWIND n.country_of_citizenship AS prop
RETURN DISTINCT prop
\end{lstlisting}\vspace{-1.6ex}
\\
\midrule
\textit{Special Categories} \\
\midrule
\raisebox{-0.6\height}{\includegraphics[height=0.4cm]{figures/match_patterns/comparison.png}} \newline\newline\texttt{Comparison} & $Q8.$ Which painting was created later, Edward George Villiers Stanley, 17th Earl of Derby or Tulip Field in Holland? \hfill (\textit{art}) & 
\vspace{-1.6ex}\begin{lstlisting}[style=TinyCypher]
MATCH (n:Painting {name: 'Edward George Villiers Stanley, 17th Earl of Derby'}), (m0:Painting {name: 'Tulip Field in Holland'})
RETURN CASE
WHEN n.creation_year > m0.creation_year THEN n.name ELSE m0.name
END AS answer
\end{lstlisting}\vspace{-1.6ex}
\\
\midrule
\raisebox{-0.6\height}{\includegraphics[height=0.4cm]{figures/match_patterns/group.png}} \newline\newline\texttt{Group By} & $Q9.$ What are the names of the mothers whose children were killed by Cersei Lannister, and how many children did Cersei kill for each of these mothers? \hfill (\textit{fictional character}) &
\vspace{-1.6ex}\begin{lstlisting}[style=TinyCypher]
MATCH (n:Character)<-[r0:hasMother]-(m0:Character)-
    [r1:killedBy]->(m1:Character {name: 'Cersei Lannister'})
WITH n, count(DISTINCT m0) AS num
RETURN n.name, num
\end{lstlisting}\vspace{-1.6ex}
\\
\midrule
\raisebox{-0.6\height}{\includegraphics[height=0.4cm]{figures/match_patterns/optional.png}} \newline\newline\texttt{Optional Match} & $Q10.$ Provide the names of all aircraft models manufactured by ATR, along with the number of flight accidents each has been involved in. \hfill (\textit{flight~accident}) & 
\vspace{-1.6ex}\begin{lstlisting}[style=TinyCypher]
MATCH (n:AircraftModel)-[r1:manufacturedBy]->
      (m1:AircraftManufacturer {name: 'ATR'}) 
OPTIONAL MATCH (n:AircraftModel)<-[r0:involves]-
               (m0:FlightAccident)
WITH n, count(DISTINCT m0) AS num
RETURN n.name, num
\end{lstlisting}\vspace{-1.6ex}
\\
\midrule
\raisebox{-0.6\height}{\includegraphics[height=0.4cm]{figures/match_patterns/time.png}}\newline\newline\texttt{Time-sensitive} & $Q11.$ Who was the CEO of Mercedes-AMG in the year 1999? \hfill (\textit{company}) & 
\vspace{-1.6ex}\begin{lstlisting}[style=TinyCypher]
MATCH (n:Person)<-[r0:hasCEO]-(m0:Company {name: 'Mercedes-AMG'})
WHERE r0.start_year <= 1999 AND (r0.end_year >= 1999 OR r0.end_year IS NULL)
WITH DISTINCT n RETURN n.name
\end{lstlisting}\vspace{-1.6ex}
\\
\midrule
\raisebox{-0.6\height}{\includegraphics[height=0.4cm]{figures/match_patterns/union.png}}\newline\newline\texttt{Union} & $Q12.$ Who are the politicians who have either led the Law and Justice party or served as the head of state of Poland at any time? \hfill (\textit{politics})  & 
\vspace{-1.6ex}\begin{lstlisting}[style=TinyCypher]
CALL {
MATCH (n:Politician)<-[r0:headedBy]-(m0:PoliticalParty {name: 'Law and Justice'}) RETURN n, m0 AS m
UNION
MATCH (n:Politician)<-[r1:hasHeadOfState]-(m1:Country {name: 'Poland'}) RETURN n, m1 AS m 
}
WITH DISTINCT n RETURN n.name
\end{lstlisting}\vspace{-1.6ex}
\\
\bottomrule
\end{tabular}
\end{adjustwidth}
\caption{ Sample questions with various graph matching patterns from the benchmark. The \textcolor[HTML]{9933cc}{nodes in purple} denote the answer entities. Square nodes \big(\raisebox{-0.1cm}{\includegraphics[height=0.4cm]{figures/match_patterns/square_gray.png}\includegraphics[height=0.4cm]{figures/match_patterns/square_color.png}}\big) denote all entities of a particular type, while circular nodes \big(\raisebox{-0.1cm}{\includegraphics[height=0.4cm]{figures/match_patterns/circle_gray.png}\includegraphics[height=0.4cm]{figures/match_patterns/circle_color.png}}\big) represent named entities. Nodes and edges with dashed lines \big(\raisebox{-0.1cm}{\includegraphics[height=0.4cm]{figures/match_patterns/dashed.png}}\big) are optional. Edges with diamond arrowheads \big(\raisebox{-0.1cm}{\includegraphics[height=0.4cm]{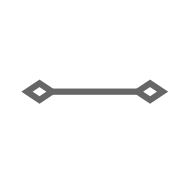}}\big) indicate relations with time sensitivity constraints. \sajhidden{Where is the definition of these concepts in the paper?}}
\label{tab:graph_patterns}
\end{table*}

\begin{table*}[!htb]
\scriptsize 
\vspace{-5pt}
\begin{adjustwidth}{-1in}{-1in}
\centering
\begin{tabular}{p{2.5cm} p{5.5cm} p{6.5cm}}
\toprule
\textbf{Return Clause Category} & \textbf{Sample Question} & \textbf{Cypher Query} \\
\midrule
\texttt{NAME} & $Q13.$ List the names of all teams. \hfill (\textit{nba})  & \vspace{-1.6ex}\begin{lstlisting}[style=TinyCypher]
MATCH (n:Team) WITH DISTINCT n
RETURN n.name
\end{lstlisting}\vspace{-1.6ex}
\\
\midrule
\texttt{PROPERTY} & $Q14.$ When did the Falcon 50 have its first flight? \hfill (\textit{flight~accident})  & \vspace{-1.6ex}\begin{lstlisting}[style=TinyCypher]
MATCH (n:AircraftModel {name: 'Falcon 50'})
WITH DISTINCT n
RETURN n.first_flight
\end{lstlisting}\vspace{-1.6ex}
\\
\midrule
\texttt{SORT} & $Q15.$ What are the names of mountains located in Nepal, sorted by elevation from lowest to highest? \hfill (\textit{geography})  & \vspace{-1.6ex}\begin{lstlisting}[style=TinyCypher]
MATCH (n:Mountain)-[r0:locatedIn]->(m0:Country {name: 'Nepal'})
WITH DISTINCT n
RETURN n.name
ORDER BY n.elevation_m ASC
\end{lstlisting}\vspace{-1.6ex}
\\
\midrule
\texttt{ARGMAX} & $Q16.$ What is the name of the Spider-Verse movie that earned the most at the global box office? \hfill (\textit{movie})  & \vspace{-1.6ex}\begin{lstlisting}[style=TinyCypher]
MATCH (n:Movie)-[r0:partOfSeries]->(m0:FilmSeries {name: 'Spider-Verse'})
WITH DISTINCT n RETURN n.name
ORDER BY n.global_box_office_usd DESC
LIMIT 1
\end{lstlisting}\vspace{-1.6ex}
\\
\midrule
\texttt{FILTER} & $Q17.$ What are the names of companies in the public relations industry that were launched after 1927? \hfill (\textit{company})  & \vspace{-1.6ex}\begin{lstlisting}[style=TinyCypher]
MATCH (n:Company)-[r0:operatesIn]->(m0:Industry {name: 'public relations'})
WITH DISTINCT n
WHERE n.launch_year > 1927
RETURN n.name
\end{lstlisting}\vspace{-1.6ex}
\\
\midrule
\texttt{AGGREGATE} & $Q18.$ What is the average longest lifespan of taxa that feed on Leporidae? \hfill (\textit{biology})  & \vspace{-1.6ex}\begin{lstlisting}[style=TinyCypher]
MATCH (n:Taxon)-[r0:feedsOn]->(m0:Taxon {name: 'Leporidae'})
WITH DISTINCT n
RETURN avg(n.longest_lifespan_years)
\end{lstlisting}\vspace{-1.6ex}
\\
\bottomrule
\end{tabular}
\end{adjustwidth}
\caption{ \texttt{RETURN} clause categories and sample questions from the benchmark. The \texttt{RETURN} clauses shown here apply to any graph pattern in the \textit{basic} categories. }
\label{tab:return_patterns}
\end{table*}

\begin{table*}[h]
\begin{adjustwidth}{-3in}{-3in}
\centering
\scriptsize
\begin{tabular}{p{3.9cm}p{3.0cm}p{4.7cm}p{4.7cm}}
\toprule
\textbf{Error Category} & \textbf{Question} & \textbf{Ground-truth Cypher} & \textbf{Predicted Cypher} \\
\midrule
\textcolor{syntaxerror}{\texttt{\textbf{Syntax Error}}} \newline
{\tiny The Cypher is not executable.}  &
What are the names and head coaches of teams that have used the Toyota Coliseum as their home venue at some point? \hfill (\textit{nba}) & 
\vspace{-1.6ex}\begin{lstlisting}[style=TinyCypher]
MATCH (n:Team)-[r0:hasHomeVenue]->
      (m0:Venue {name: 'Toyota Coliseum'})
WITH DISTINCT n
RETURN n.name, n.head_coach
\end{lstlisting}\vspace{-1.6ex} & 
\vspace{-1.6ex}\begin{lstlisting}[style=TinyCypher]
MATCH (t:Team)-
[:hasHomeVenue {start_year: start, end_year: end}]->
(v:Venue {name: 'Toyota Coliseum'})
RETURN DISTINCT t.name, t.head_coach
\end{lstlisting}\vspace{-1.6ex} \\
\midrule
\textcolor{wrongentrelpropertytype}{\texttt{\textbf{Wrong Ent/Rel/Property type}}} \newline
{\tiny The Cypher uses a valid but not intended entity/relation/property type.}  &
What is the name of the movie from the same country where The Fighter was released, which has the highest global box office earnings? \hfill (\textit{movie}) & 
\vspace{-1.6ex}\begin{lstlisting}[style=TinyCypher]
MATCH (n:Movie)-
      [r0:originatesFrom]->
      (m0:Country)<-[r1:releasedIn]-
      (m1:Movie {name: 'The Fighter'})
WITH DISTINCT n RETURN n.name
ORDER BY n.global_box_office_usd DESC LIMIT 1
\end{lstlisting}\vspace{-1.6ex} & 
\vspace{-1.6ex}\begin{lstlisting}[style=TinyCypher]
MATCH (m:Movie)-[:releasedIn]->
      (c:Country)<-[:releasedIn]-
      (f:Movie {name: 'The Fighter'})
RETURN m.name
ORDER BY m.global_box_office_usd DESC LIMIT 1
\end{lstlisting}\vspace{-1.6ex} \\
\midrule
\textcolor{reverseddirection}{\texttt{\textbf{Reversed Direction}}} \newline
{\tiny A relation is used in the opposite direction.}  &
What is the name of the subsidiary of The Coca-Cola Company that was launched most recently? \hfill (\textit{company}) & 
\vspace{-1.6ex}\begin{lstlisting}[style=TinyCypher]
MATCH (n:Company)-[r0:subsidiaryOf]->
    (m0:Company {name: 'The Coca-Cola Company'})
WITH DISTINCT n RETURN n.name
ORDER BY n.launch_year DESC LIMIT 1
\end{lstlisting}\vspace{-1.6ex} & 
\vspace{-1.6ex}\begin{lstlisting}[style=TinyCypher]
MATCH (parent:Company {name: 'The Coca-Cola Company'})
  -[:subsidiaryOf]->(subsidiary:Company)
RETURN subsidiary.name ORDER BY
  subsidiary.launch_year DESC LIMIT 1
\end{lstlisting}\vspace{-1.6ex} \\
\midrule
\textcolor{entitylinking}{\texttt{\textbf{Entity Linking}}} \newline
{\tiny An entity name in the Cypher does not match the name of the intended entity in the database.}  &
Which ocean, the Atlantic or the Arctic, has a greater maximum vertical depth? \hfill (\textit{geography}) & 
\vspace{-1.6ex}\begin{lstlisting}[style=TinyCypher]
MATCH (n:Ocean {name: 'Atlantic Ocean'}),
    (m0:Ocean {name: 'Arctic Ocean'})
RETURN CASE WHEN n.max_vertical_depth_m > m0.max_vertical_depth_m THEN n.name ELSE m0.name END AS answer
\end{lstlisting}\vspace{-1.6ex} & 
\vspace{-1.6ex}\begin{lstlisting}[style=TinyCypher]
MATCH (o:Ocean) WHERE o.name IN ['Atlantic', 'Arctic']
RETURN o.name
ORDER BY o.max_vertical_depth_m DESC LIMIT 1
\end{lstlisting}\vspace{-1.6ex} \\
\midrule
\textcolor{schemaviolation}{\texttt{\textbf{Schema violation}}}\newline {\tiny The MATCH pattern does not conform to the graph schema (aside from reversed relations). For example, the Cypher contains a hallucinated entity/relation/property type, or a relation that is invalid between its subject and object.} & 
What are the names of film series that include movies originating from the Czech Republic? \hfill (\textit{movie}) & 
\vspace{-1.6ex}\begin{lstlisting}[style=TinyCypher]
MATCH (n:FilmSeries)<-
      [r0:partOfSeries]-(m0:Movie)-
      [r1:originatesFrom]->
(m1:Country {name: 'Czech Republic'}) 
WITH DISTINCT n 
RETURN n.name
\end{lstlisting}\vspace{-1.6ex} & 
\vspace{-1.6ex}\begin{lstlisting}[style=TinyCypher]
MATCH (m:Movie)-[:originatesFrom]->
      (c:Country {name: 'Czech Republic'})-
      [:partOfSeries]->(fs:FilmSeries) 
RETURN DISTINCT fs.name
\end{lstlisting}\vspace{-1.6ex} \\
\midrule
\textcolor{patternnotalignedwithquestion}{\texttt{\textbf{Pattern Not Aligned with Question}}} \newline {\tiny The MATCH pattern conforms to the graph schema, but does not align with the question.} & 
What are the names of aviation accidents that either departed from Aeroparque Jorge Newbery or were headed to Ingeniero Aeronáutico Ambrosio L.V. Taravella International Airport? \hfill (\textit{flight accident}) & 
\vspace{-1.6ex}\begin{lstlisting}[style=TinyCypher]
CALL { MATCH (n:FlightAccident)-
  [r0:departsFrom]->(m0:Airport
  {name: 'Aeroparque Jorge Newbery'}) RETURN n, m0 as m
  UNION MATCH (n:FlightAccident)-
  [r1:destinedFor]->(m1:Airport
  {name: 'Ingeniero Aeronáutico Ambrosio L.V. Taravella International Airport'})
  RETURN n, m1 as m } 
WITH DISTINCT n RETURN n.name
\end{lstlisting}\vspace{-1.6ex} & 
\vspace{-1.6ex}\begin{lstlisting}[style=TinyCypher]
MATCH (fa:FlightAccident)-
      [:departsFrom|:destinedFor]->
      (a:Airport)
WHERE a.name IN ['Aeroparque Jorge Newbery', 'Ingeniero Aeronáutico Ambrosio L.V. Taravella International Airport']
RETURN DISTINCT fa.name
\end{lstlisting}\vspace{-1.6ex} \\
\midrule
\textcolor{incorrectgrouping}{\texttt{\textbf{Incorrect Grouping}}} \newline {\tiny The grouping operation in the Cypher is incorrect.} & 
Provide the names of all lakes that are part of the Congo basin, along with the number of countries each lake is located in. \hfill (\textit{geography}) & 
\vspace{-1.6ex}\begin{lstlisting}[style=TinyCypher]
MATCH (n:Lake)-[r1:partOf]->
      (m1:DrainageBasin {name: 'Congo basin'})
OPTIONAL MATCH (n:Lake)-
   [r0:locatedIn]->(m0:Country)
WITH n, count(DISTINCT m0) AS num
RETURN n.name, num
\end{lstlisting}\vspace{-1.6ex} & 
\vspace{-1.6ex}\begin{lstlisting}[style=TinyCypher]
MATCH (l:Lake)-[:partOf]->
      (d:DrainageBasin {name: 'Congo'}),
      (l)-[:locatedIn]->(c:Country)
RETURN l.name AS lakeName, count(DISTINCT c) AS numberOfCountries
\end{lstlisting}\vspace{-1.6ex} \\
\midrule
\textcolor{incorrectfiltering}{\texttt{\textbf{Incorrect Filtering}}} \newline {\tiny The filtering operation in the Cypher is incorrect.} & 
What are the names of lakes situated in countries through which the Natara River flows, with an area of less than 390,000 square kilometers? \hfill (\textit{geography}) & 
\vspace{-1.6ex}\begin{lstlisting}[style=TinyCypher]
MATCH (n:Lake)-[r0:locatedIn]->
      (m0:Country)<-[r1:flowsThrough]-
      (m1:River {name: 'Natara'}) 
WITH DISTINCT n 
WHERE n.area_km2 < 390000 
RETURN n.name
\end{lstlisting}\vspace{-1.6ex} & 
\vspace{-1.6ex}\begin{lstlisting}[style=TinyCypher]
MATCH (r:River {name: 'Natara'})-
    [:flowsThrough]->(c:Country),
    (c)-[:locatedIn]->(co:Continent),
    (r)-[:flowsInto]->(l:Lake)
WHERE c.area_km2 < 390000
RETURN DISTINCT l.name
\end{lstlisting}\vspace{-1.6ex} \\
\midrule
\textcolor{resultsnotalignedwithquestion}{\texttt{\textbf{Results Not Aligned with Question}}}
\newline {\tiny The RETURN clause does not align with the question, resulting in outputs that do not match the intended results of the query.} & 
What are the unique countries of citizenship for characters who have a father? \hfill (\textit{fictional character}) & 
\vspace{-1.6ex}\begin{lstlisting}[style=TinyCypher]
MATCH (n:Character)-[r0:hasFather]->
    (m0:Character)
WITH DISTINCT n UNWIND
  n.country_of_citizenship AS prop
RETURN DISTINCT prop
\end{lstlisting}\vspace{-1.6ex} & 
\vspace{-1.6ex}\begin{lstlisting}[style=TinyCypher]
MATCH (c:Character)-[:hasFather]->
    (:Character)
RETURN DISTINCT c.country_of_citizenship AS countries
\end{lstlisting}\vspace{-1.6ex} \\
\midrule
\textcolor{incorrectdeduplication}{\texttt{\textbf{Incorrect Deduplication}}} \newline {\tiny The Cypher execution results contain the same set of entries as the ground truth, but include entries that should be deduplicated, or missing entries that should be duplicated.} & 
What are the names of mountains that are part of the Andes mountain range? \hfill (\textit{geography}) & 
\vspace{-1.6ex}\begin{lstlisting}[style=TinyCypher]
MATCH (n:Mountain)-[r0:partOf]->
      (m0:MountainRange {name: 'Andes'})
WITH DISTINCT n 
RETURN n.name
\end{lstlisting}\vspace{-1.6ex} & 
\vspace{-1.6ex}\begin{lstlisting}[style=TinyCypher]
MATCH (m:Mountain)-[:partOf]->
      (mr:MountainRange {name: 'Andes'})
RETURN DISTINCT m.name
\end{lstlisting}\vspace{-1.6ex} \\
\bottomrule
\end{tabular}
\end{adjustwidth}
\caption{Definitions and examples for the 10 text-to-Cypher error categories.}
\label{tab:error_definition}
\end{table*}

\begin{table*}[!htb]
\scriptsize 
\vspace{-5pt}
\begin{adjustwidth}{-2in}{-2in}
\centering
\begin{tabular}{p{1.8cm} >{\centering\arraybackslash}m{4.3cm} p{10cm}}
\toprule
 \textbf{Name} & \raggedright\arraybackslash \textbf{Graph Schema} & \textbf{Entity/Relation Properties} \\
\midrule
\textit{art} & \raisebox{-0.9 \height}{\includegraphics[width=4.2cm]{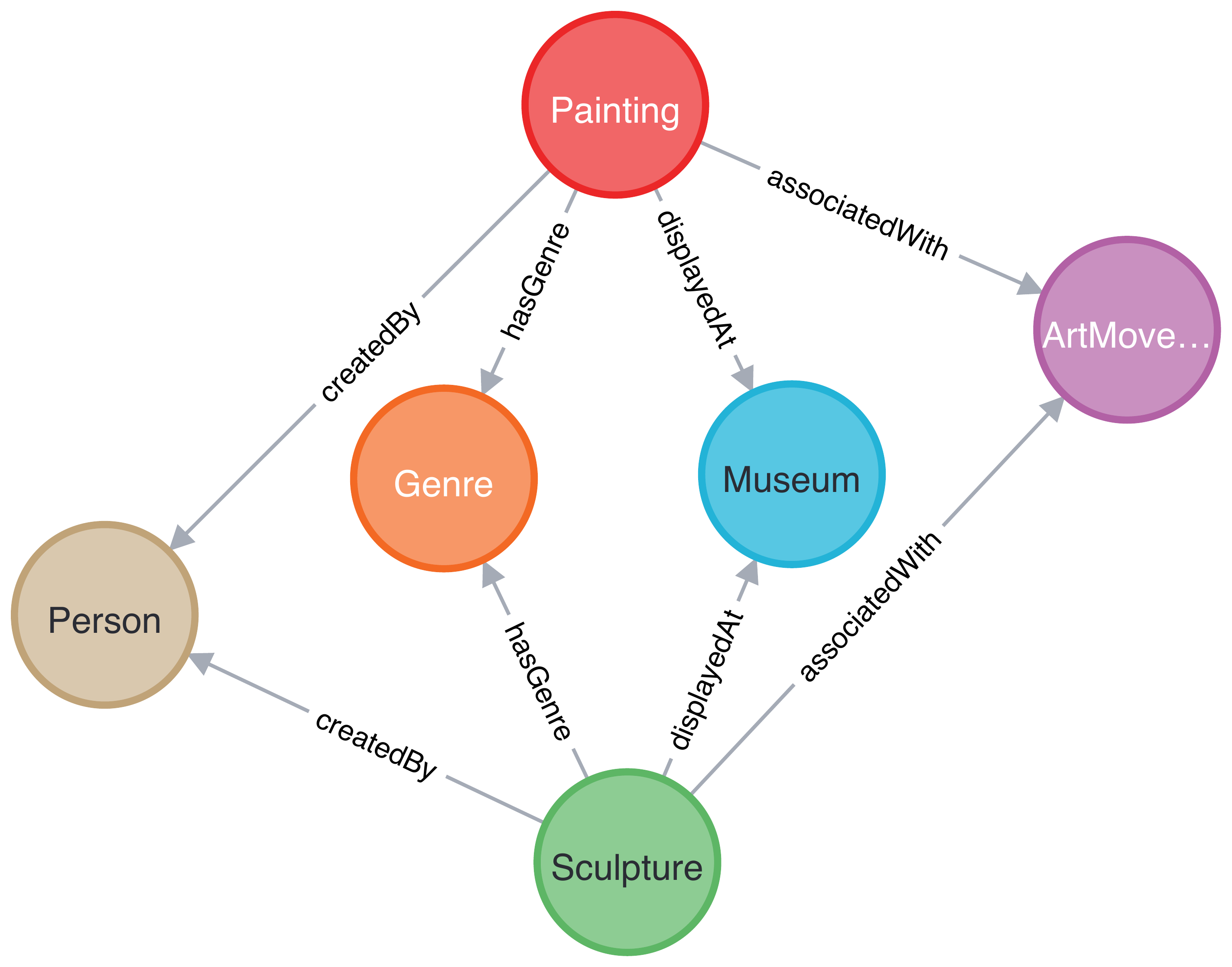}} &
\RaggedRight \tiny
\nlabel{Painting}  \nprop{name \textcolor{white}{STR}}  \nprop{creation\_year \textcolor{white}{INT}}  \nprop{country\_of\_origin \textcolor{white}{STR}} \quad  \nlabel{ArtMovement}  \nprop{start\_year \textcolor{white}{INT}}  \nprop{name \textcolor{white}{STR}}  \nprop{end\_year \textcolor{white}{INT}} \quad  \nlabel{Sculpture}  \nprop{name \textcolor{white}{STR}}  \nprop{creation\_year \textcolor{white}{INT}}  \nprop{country\_of\_origin \textcolor{white}{STR}} \quad  \nlabel{Person}  \nprop{place\_of\_birth \textcolor{white}{STR}}  \nprop{name \textcolor{white}{STR}}  \nprop{gender \textcolor{white}{STR}}  \nprop{date\_of\_death \textcolor{white}{DATE}}  \nprop{date\_of\_birth \textcolor{white}{DATE}} \quad  \nlabel{Museum}  \nprop{name \textcolor{white}{STR}} \quad  \nlabel{Genre}  \nprop{name \textcolor{white}{STR}}
\\
\midrule
\textit{biology} & \raisebox{-0.9 \height}{\includegraphics[width=3.8cm]{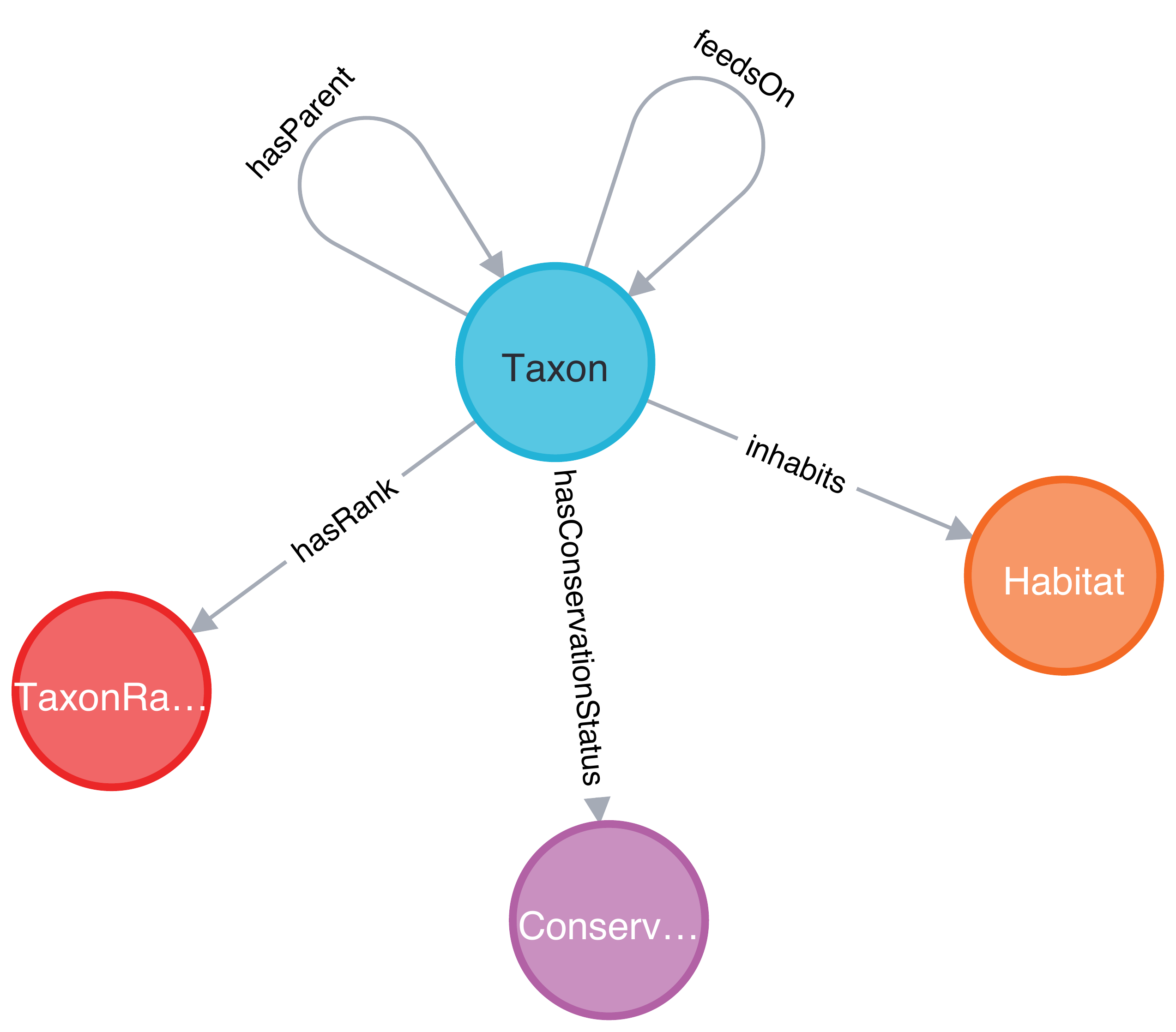}} &
\RaggedRight \tiny
\nlabel{Taxon}  \nprop{taxon\_name \textcolor{white}{STR}}  \nprop{name \textcolor{white}{STR}}  \nprop{longest\_lifespan\_years \textcolor{white}{FLOAT}}  \nprop{diel\_cycle \textcolor{white}{STR}}  \nprop{avg\_gestation\_period\_days \textcolor{white}{FLOAT}} \quad  \nlabel{ConservationStatus}  \nprop{name \textcolor{white}{STR}} \quad  \nlabel{TaxonRank}  \nprop{name \textcolor{white}{STR}} \quad  \nlabel{Habitat}  \nprop{name \textcolor{white}{STR}}
\\
\midrule
\textit{company} & \raisebox{-0.9 \height}{\includegraphics[width=3.8cm]{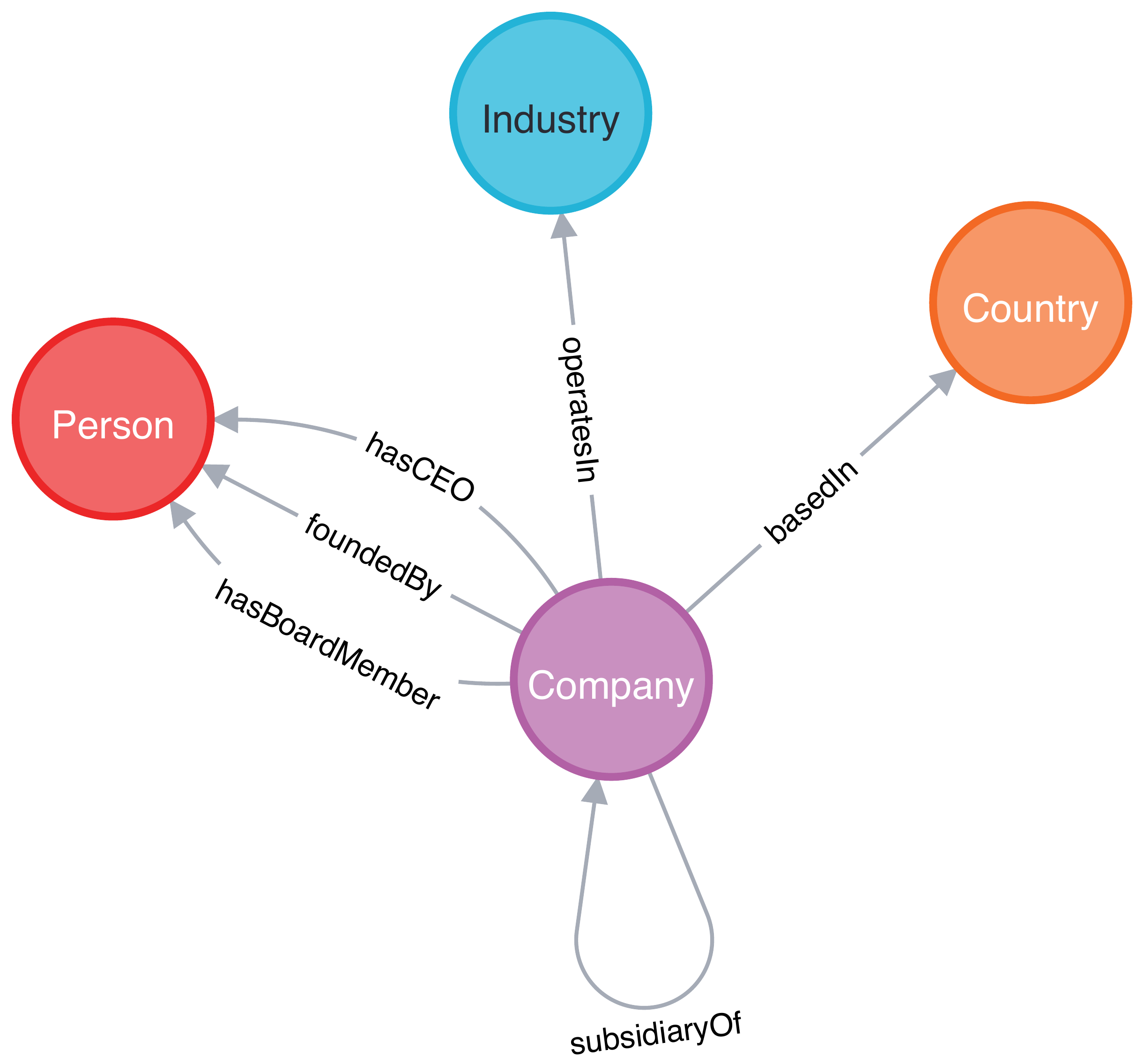}} &
\RaggedRight \tiny
\nlabel{Company}  \nprop{name \textcolor{white}{STR}}  \nprop{launch\_year \textcolor{white}{INT}} \quad  \nlabel{Country}  \nprop{name \textcolor{white}{STR}} \quad  \nlabel{Person}  \nprop{place\_of\_birth \textcolor{white}{STR}}  \nprop{name \textcolor{white}{STR}}  \nprop{gender \textcolor{white}{STR}}  \nprop{date\_of\_death \textcolor{white}{DATE}}  \nprop{date\_of\_birth \textcolor{white}{DATE}}  \nprop{country\_of\_citizenship \textcolor{white}{LIST[STR]}} \quad  \nlabel{Industry}  \nprop{name \textcolor{white}{STR}} \newline   \rlabel{hasBoardMember}  \rprop{start\_year \textcolor{white}{INT}}  \rprop{end\_year \textcolor{white}{INT}} \quad  \rlabel{hasCEO}  \rprop{start\_year \textcolor{white}{INT}}  \rprop{end\_year \textcolor{white}{INT}}
\\
\midrule
\textit{fictional~character} & \raisebox{-0.9 \height}{\includegraphics[width=4cm]{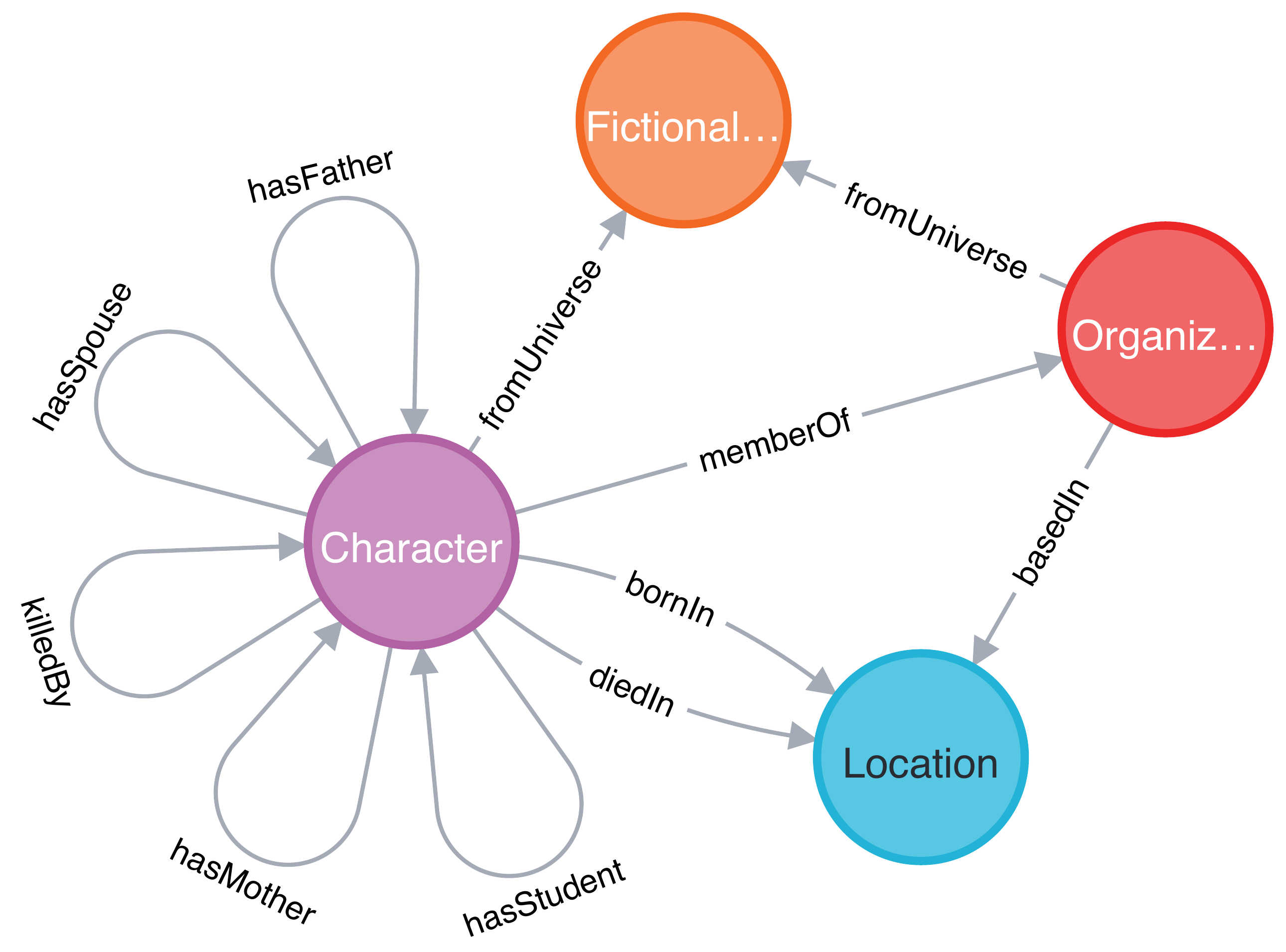}} &
\RaggedRight \tiny
\nlabel{Organization}  \nprop{name \textcolor{white}{STR}} \quad  \nlabel{Location}  \nprop{name \textcolor{white}{STR}} \quad  \nlabel{Character}  \nprop{occupation \textcolor{white}{LIST[STR]}}  \nprop{name \textcolor{white}{STR}}  \nprop{gender \textcolor{white}{STR}}  \nprop{creator \textcolor{white}{STR}}  \nprop{country\_of\_citizenship \textcolor{white}{LIST[STR]}}  \nprop{birth\_name \textcolor{white}{STR}} \quad  \nlabel{FictionalUniverse}  \nprop{name \textcolor{white}{STR}}  \nprop{inception\_year \textcolor{white}{INT}}  \nprop{creator \textcolor{white}{STR}}
\\
 \midrule
\textit{flight accident} & \raisebox{-0.9 \height}{\includegraphics[width=4.2cm]{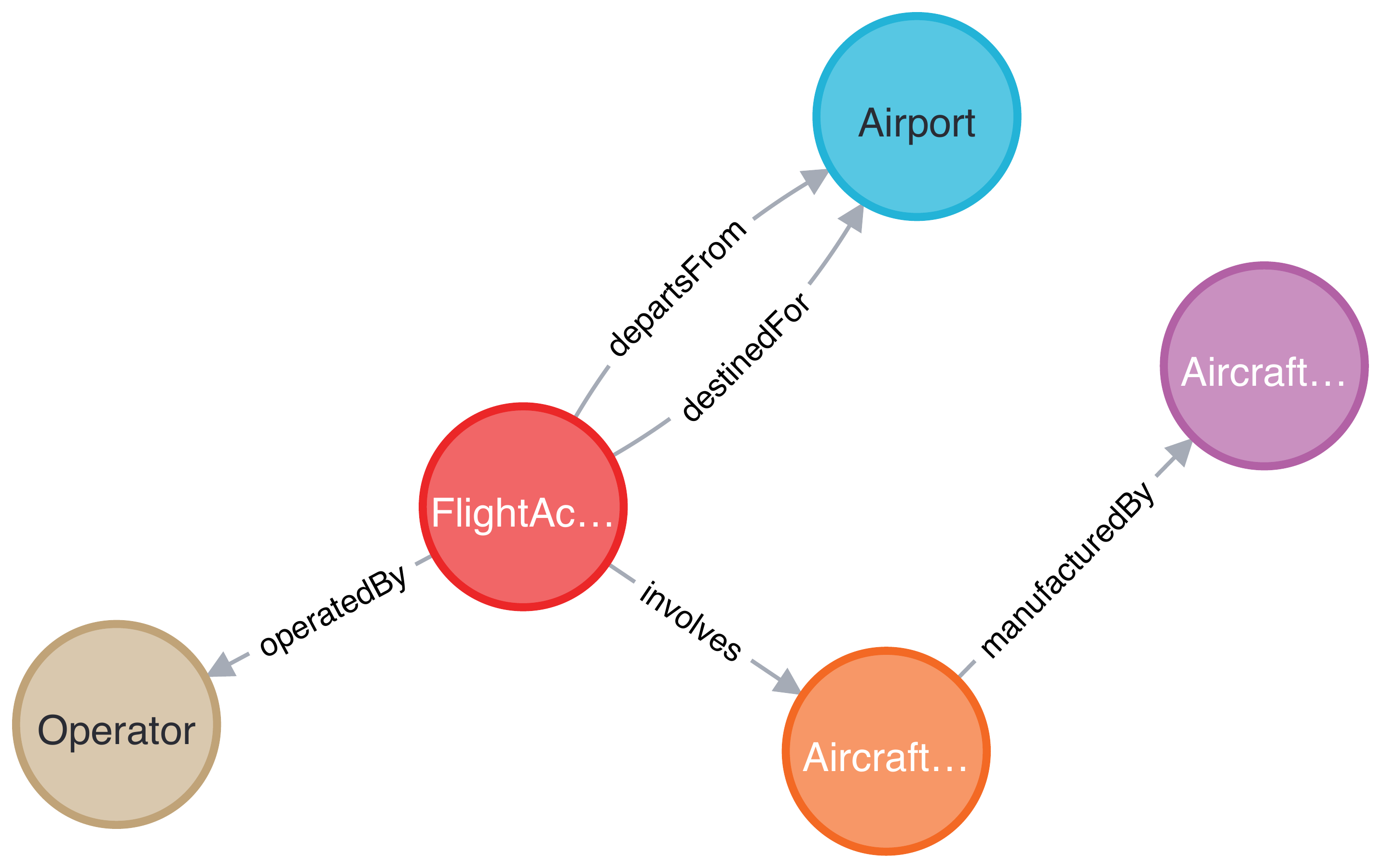}} &
\RaggedRight \tiny
\nlabel{FlightAccident}  \nprop{number\_of\_survivors \textcolor{white}{INT}}  \nprop{number\_of\_injuries \textcolor{white}{INT}}  \nprop{number\_of\_deaths \textcolor{white}{INT}}  \nprop{name \textcolor{white}{STR}}  \nprop{location \textcolor{white}{STR}}  \nprop{flight\_number \textcolor{white}{STR}}  \nprop{date \textcolor{white}{DATE}} \quad  \nlabel{Airport}  \nprop{name \textcolor{white}{STR}}  \nprop{location \textcolor{white}{STR}}  \nprop{icao\_code \textcolor{white}{STR}}  \nprop{iata\_code \textcolor{white}{STR}}  \nprop{country \textcolor{white}{STR}} \quad  \nlabel{AircraftModel}  \nprop{wingspan\_metre \textcolor{white}{FLOAT}}  \nprop{service\_entry \textcolor{white}{DATE}}  \nprop{range\_km \textcolor{white}{FLOAT}}  \nprop{name \textcolor{white}{STR}}  \nprop{length\_metre \textcolor{white}{FLOAT}}  \nprop{height\_metre \textcolor{white}{FLOAT}}  \nprop{first\_flight \textcolor{white}{DATE}} \quad  \nlabel{AircraftManufacturer}  \nprop{name \textcolor{white}{STR}}  \nprop{launch\_year \textcolor{white}{INT}}  \nprop{country \textcolor{white}{STR}} \quad  \nlabel{Operator}  \nprop{name \textcolor{white}{STR}}  \nprop{launch\_year \textcolor{white}{INT}}  \nprop{country \textcolor{white}{STR}} \newline
\\
\midrule
\textit{geography} & \raisebox{-0.9 \height}{\includegraphics[width=4.3cm]{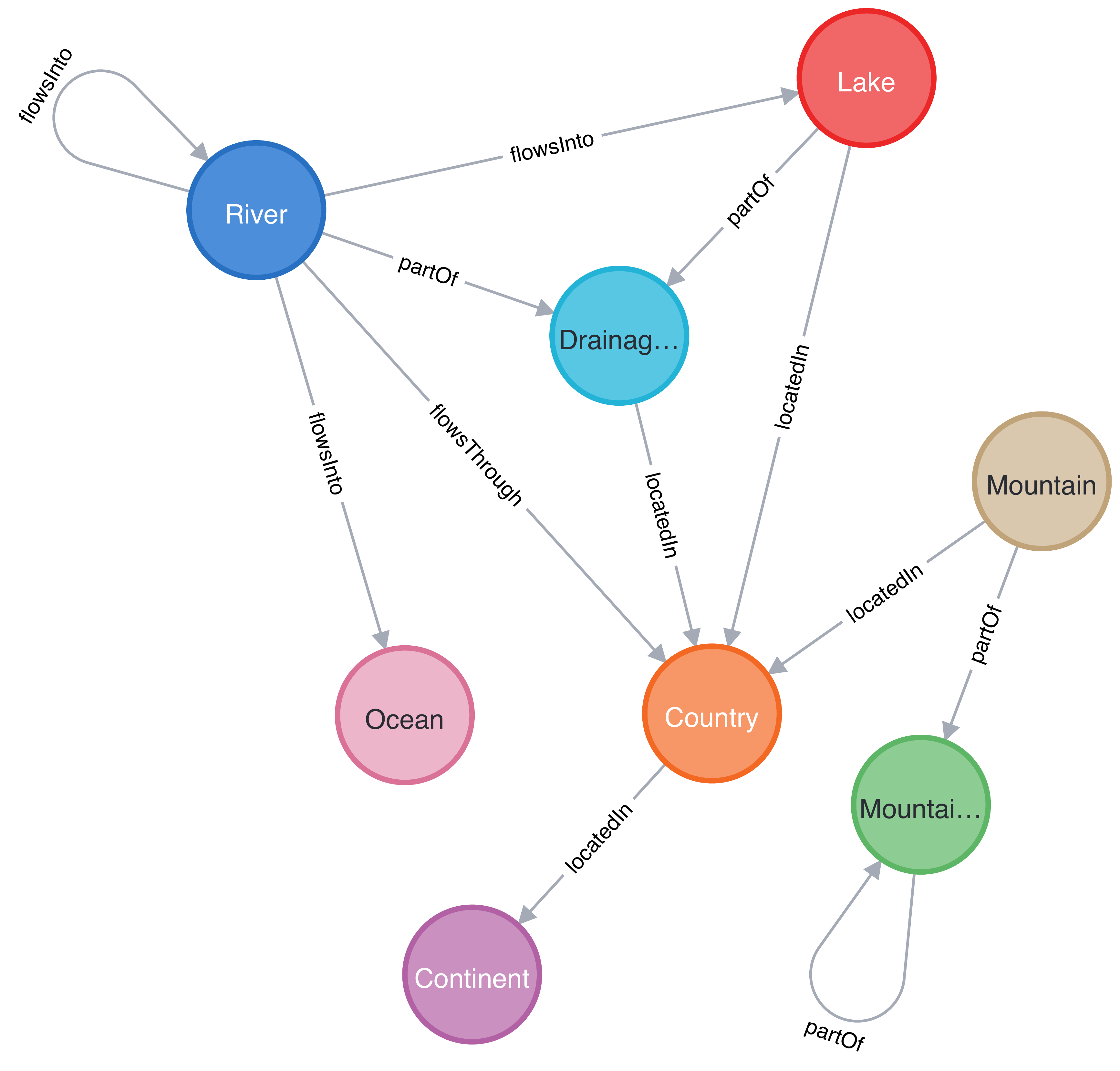}} &
\RaggedRight \tiny
\nlabel{River}  \nprop{name \textcolor{white}{STR}}  \nprop{length\_km \textcolor{white}{FLOAT}}  \nprop{discharge\_m3\_s \textcolor{white}{FLOAT}} \quad  \nlabel{Lake}  \nprop{vertical\_depth\_m \textcolor{white}{FLOAT}}  \nprop{name \textcolor{white}{STR}}  \nprop{area\_km2 \textcolor{white}{FLOAT}} \quad  \nlabel{Ocean}  \nprop{name \textcolor{white}{STR}}  \nprop{max\_vertical\_depth\_m \textcolor{white}{FLOAT}}  \nprop{avg\_vertical\_depth\_m \textcolor{white}{FLOAT}}  \nprop{area\_km2 \textcolor{white}{FLOAT}} \quad  \nlabel{Country}  \nprop{name \textcolor{white}{STR}}  \nprop{capital \textcolor{white}{STR}}  \nprop{area\_km2 \textcolor{white}{FLOAT}} \quad  \nlabel{Continent}  \nprop{name \textcolor{white}{STR}} \quad  \nlabel{DrainageBasin}  \nprop{name \textcolor{white}{STR}}  \nprop{area\_km2 \textcolor{white}{FLOAT}} \quad  \nlabel{Mountain}  \nprop{name \textcolor{white}{STR}}  \nprop{elevation\_m \textcolor{white}{FLOAT}} \quad  \nlabel{MountainRange}  \nprop{name \textcolor{white}{STR}}
\\
\bottomrule
\end{tabular}
\end{adjustwidth}
\caption{ Schemas of the 11 graphs in the benchmark. The color of the property boxes indicates whether they are \textcolor{beige}{entity properties} \big(\eg \raisebox{6pt}{\scriptsize \nprop{name \textcolor{white}{STR}}}\big) or \textcolor{olive}{relation properties} \big(\eg \raisebox{6pt}{\scriptsize \rprop{start\_year \textcolor{white}{INT}}}\big).  }
\label{tab:all_graphs}
\end{table*}

\begin{table*}[!htb]
\scriptsize 
\vspace{-5pt}
\begin{adjustwidth}{-2in}{-2in}
\centering
\begin{tabular}{p{1.8cm} >{\centering\arraybackslash}m{4.3cm} p{10cm}}
\toprule
 \textbf{Name} & \raggedright\arraybackslash \textbf{Graph Schema} & \textbf{Entity/Relation Properties} \\

\midrule
\textit{movie} & \raisebox{-0.9 \height}{\includegraphics[width=4cm]{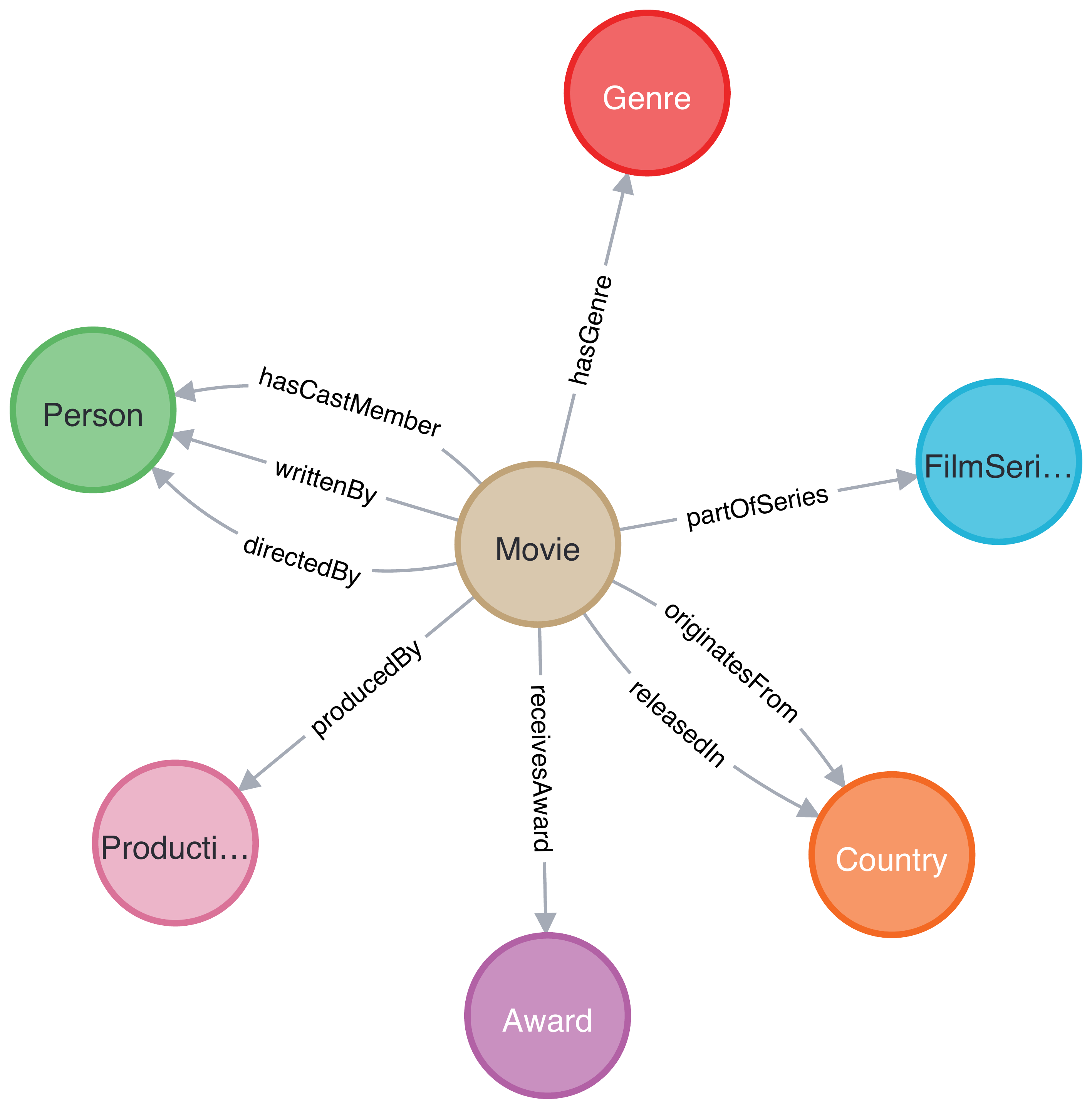}} &
\RaggedRight \tiny
\nlabel{Movie}  \nprop{runtime\_minute \textcolor{white}{FLOAT}}  \nprop{original\_language \textcolor{white}{LIST[STR]}}  \nprop{name \textcolor{white}{STR}}  \nprop{global\_box\_office\_usd \textcolor{white}{FLOAT}}  \nprop{filming\_location \textcolor{white}{LIST[STR]}} \quad  \nlabel{Person}  \nprop{place\_of\_birth \textcolor{white}{STR}}  \nprop{name \textcolor{white}{STR}}  \nprop{gender \textcolor{white}{STR}}  \nprop{date\_of\_death \textcolor{white}{DATE}}  \nprop{date\_of\_birth \textcolor{white}{DATE}}  \nprop{country\_of\_citizenship \textcolor{white}{LIST[STR]}} \quad  \nlabel{Genre}  \nprop{name \textcolor{white}{STR}} \quad  \nlabel{Country}  \nprop{name \textcolor{white}{STR}} \quad  \nlabel{FilmSeries}  \nprop{name \textcolor{white}{STR}} \quad  \nlabel{ProductionCompany}  \nprop{name \textcolor{white}{STR}}  \nprop{country \textcolor{white}{STR}} \quad  \nlabel{Award}  \nprop{name \textcolor{white}{STR}} \newline   \rlabel{hasCastMember}  \rprop{character\_role \textcolor{white}{STR}} \quad  \rlabel{receivesAward}  \rprop{year \textcolor{white}{INT}}  \rprop{winners \textcolor{white}{LIST[STR]}} \quad  \rlabel{releasedIn}  \rprop{date \textcolor{white}{DATE}}
\\
\midrule
\textit{nba} & \raisebox{-0.9 \height}{\includegraphics[width=4cm]{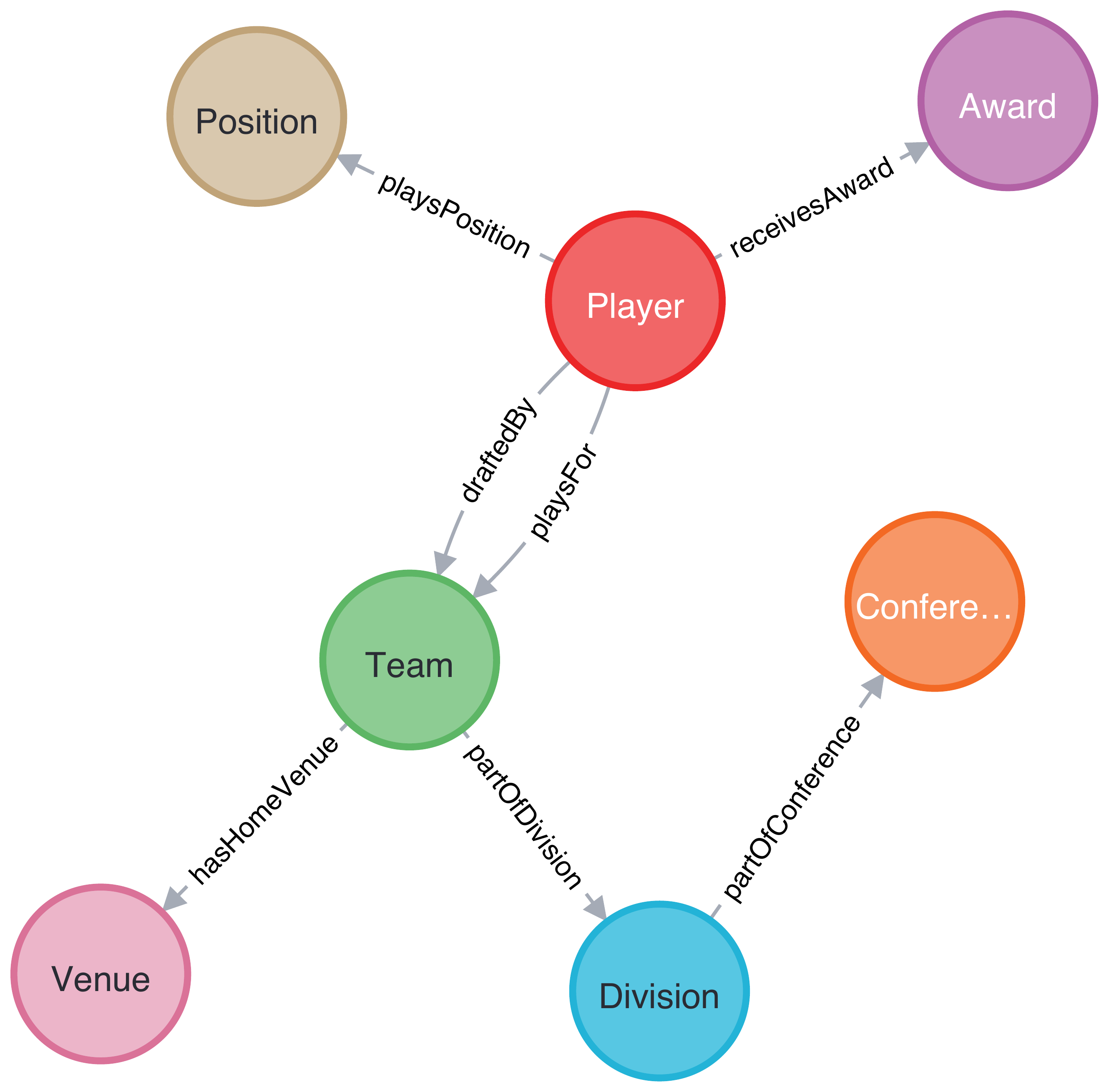}} &
\RaggedRight \tiny
 \nlabel{Player}  \nprop{schools\_attended \textcolor{white}{LIST[STR]}}  \nprop{place\_of\_birth \textcolor{white}{STR}}  \nprop{nicknames \textcolor{white}{LIST[STR]}}  \nprop{name \textcolor{white}{STR}}  \nprop{mass\_kg \textcolor{white}{FLOAT}}  \nprop{height\_cm \textcolor{white}{FLOAT}}  \nprop{handedness \textcolor{white}{STR}}  \nprop{gender \textcolor{white}{STR}}  \nprop{date\_of\_death \textcolor{white}{DATE}}  \nprop{date\_of\_birth \textcolor{white}{DATE}}  \nprop{country\_of\_citizenship \textcolor{white}{LIST[STR]}} \quad  \nlabel{Team}  \nprop{owners \textcolor{white}{LIST[STR]}}  \nprop{name \textcolor{white}{STR}}  \nprop{inception\_year \textcolor{white}{INT}}  \nprop{head\_coach \textcolor{white}{STR}} \quad  \nlabel{Venue}  \nprop{name \textcolor{white}{STR}} \quad  \nlabel{Division}  \nprop{name \textcolor{white}{STR}} \quad  \nlabel{Conference}  \nprop{name \textcolor{white}{STR}} \quad  \nlabel{Position}  \nprop{name \textcolor{white}{STR}} \quad  \nlabel{Award}  \nprop{name \textcolor{white}{STR}} \newline   \rlabel{draftedBy}  \rprop{year \textcolor{white}{INT}} \quad  \rlabel{hasHomeVenue}  \rprop{start\_year \textcolor{white}{INT}}  \rprop{end\_year \textcolor{white}{INT}} \quad  \rlabel{playsFor}  \rprop{start\_year \textcolor{white}{INT}}  \rprop{sport\_number \textcolor{white}{INT}}  \rprop{end\_year \textcolor{white}{INT}} \quad  \rlabel{receivesAward}  \rprop{year \textcolor{white}{INT}}
\\
 \midrule
\textit{politics} & \raisebox{-0.9 \height}{\includegraphics[width=4.3cm]{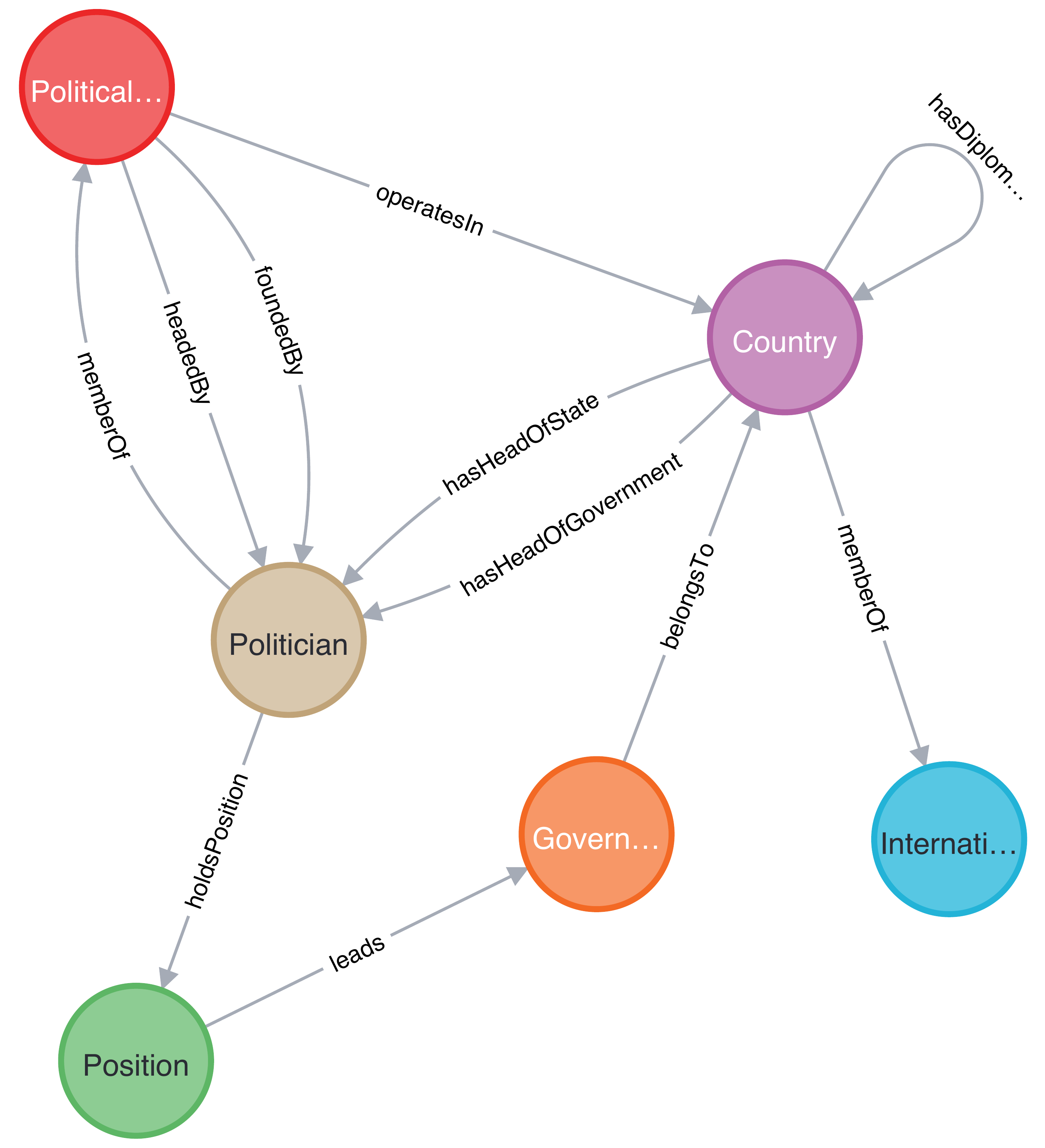}} &
\RaggedRight \tiny
\nlabel{GovernmentOrganization}  \nprop{name \textcolor{white}{STR}} \quad  \nlabel{Country}  \nprop{official\_language \textcolor{white}{LIST[STR]}}  \nprop{name \textcolor{white}{STR}}  \nprop{founding\_date \textcolor{white}{DATE}} \quad  \nlabel{PoliticalParty}  \nprop{name \textcolor{white}{STR}}  \nprop{founding\_date \textcolor{white}{DATE}} \quad  \nlabel{Politician}  \nprop{schools\_attended \textcolor{white}{LIST[STR]}}  \nprop{place\_of\_death \textcolor{white}{STR}}  \nprop{place\_of\_birth \textcolor{white}{STR}}  \nprop{name \textcolor{white}{STR}}  \nprop{gender \textcolor{white}{STR}}  \nprop{date\_of\_death \textcolor{white}{DATE}}  \nprop{date\_of\_birth \textcolor{white}{DATE}}  \nprop{country\_of\_citizenship \textcolor{white}{LIST[STR]}} \quad  \nlabel{Position}  \nprop{name \textcolor{white}{STR}} \quad  \nlabel{InternationalOrganization}  \nprop{name \textcolor{white}{STR}}  \nprop{founding\_year \textcolor{white}{INT}} \newline   \rlabel{hasHeadOfGovernment}  \rprop{start\_year \textcolor{white}{INT}}  \rprop{end\_year \textcolor{white}{INT}} \quad  \rlabel{hasHeadOfState}  \rprop{start\_year \textcolor{white}{INT}}  \rprop{end\_year \textcolor{white}{INT}} \quad  \rlabel{headedBy}  \rprop{start\_year \textcolor{white}{INT}}  \rprop{end\_year \textcolor{white}{INT}} \quad  \rlabel{holdsPosition}  \rprop{start\_year \textcolor{white}{INT}}  \rprop{end\_year \textcolor{white}{INT}}
\\
\midrule
\textit{soccer} & \raisebox{-0.9 \height}{\includegraphics[width=3.8cm]{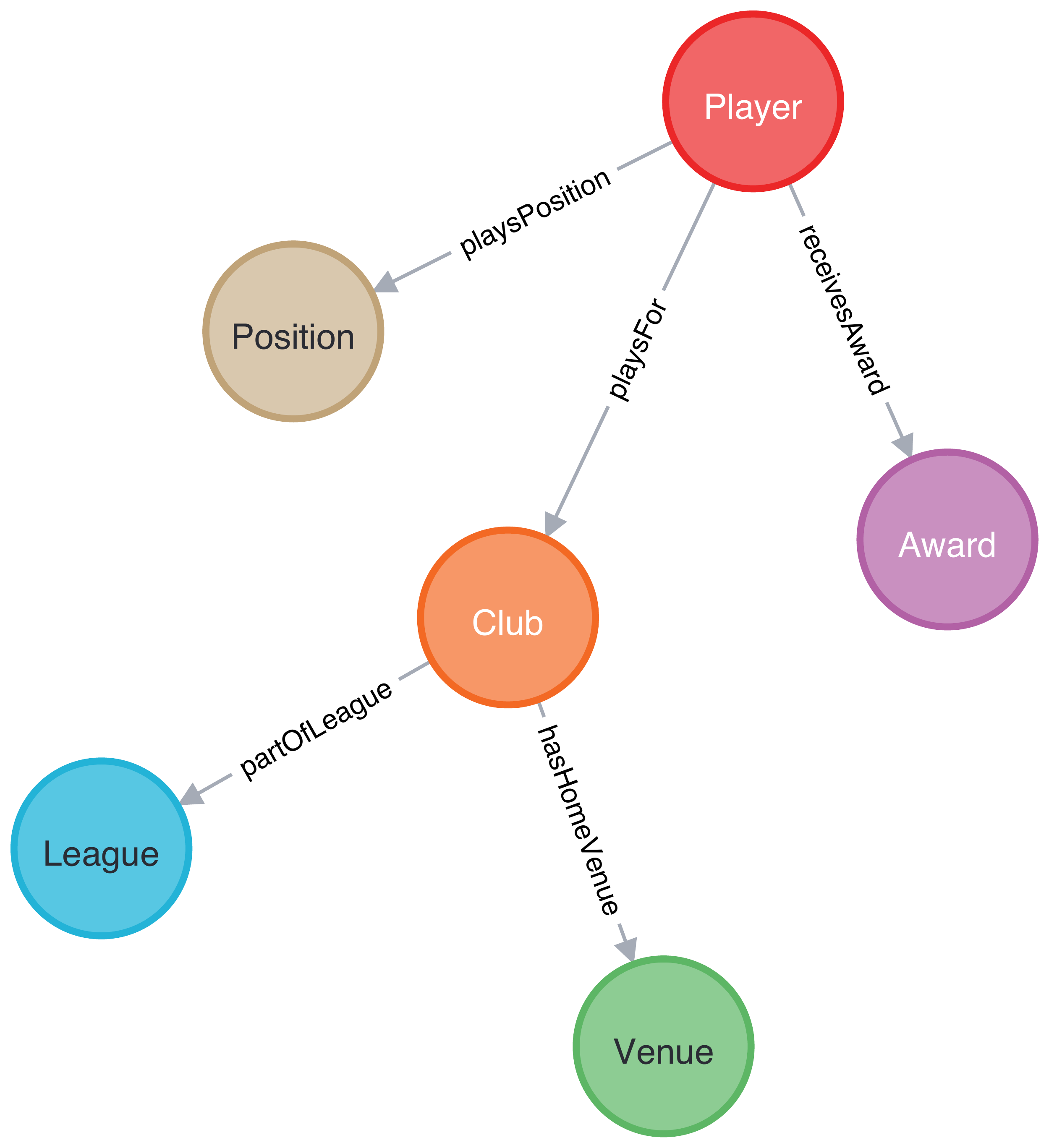}} &
\RaggedRight \tiny
\nlabel{Club}  \nprop{owners \textcolor{white}{LIST[STR]}}  \nprop{name \textcolor{white}{STR}}  \nprop{inception\_year \textcolor{white}{INT}}  \nprop{head\_coach \textcolor{white}{STR}}  \nprop{country \textcolor{white}{STR}} \quad  \nlabel{Venue}  \nprop{name \textcolor{white}{STR}} \quad  \nlabel{League}  \nprop{name \textcolor{white}{STR}} \quad  \nlabel{Player}  \nprop{schools\_attended \textcolor{white}{LIST[STR]}}  \nprop{place\_of\_birth \textcolor{white}{STR}}  \nprop{nicknames \textcolor{white}{LIST[STR]}}  \nprop{name \textcolor{white}{STR}}  \nprop{mass\_kg \textcolor{white}{FLOAT}}  \nprop{height\_cm \textcolor{white}{FLOAT}}  \nprop{gender \textcolor{white}{STR}}  \nprop{footedness \textcolor{white}{STR}}  \nprop{date\_of\_death \textcolor{white}{DATE}}  \nprop{date\_of\_birth \textcolor{white}{DATE}}  \nprop{country\_of\_citizenship \textcolor{white}{LIST[STR]}} \quad  \nlabel{Position}  \nprop{name \textcolor{white}{STR}} \quad  \nlabel{Award}  \nprop{name \textcolor{white}{STR}} \newline   \rlabel{hasHomeVenue}  \rprop{start\_year \textcolor{white}{INT}}  \rprop{end\_year \textcolor{white}{INT}} \quad  \rlabel{playsFor}  \rprop{start\_year \textcolor{white}{INT}}  \rprop{sport\_number \textcolor{white}{INT}}  \rprop{end\_year \textcolor{white}{INT}} \quad  \rlabel{receivesAward}  \rprop{year \textcolor{white}{INT}}\newline
\\
\midrule
\textit{terrorist attack} & \raisebox{-0.9 \height}{\includegraphics[width=3.8cm]{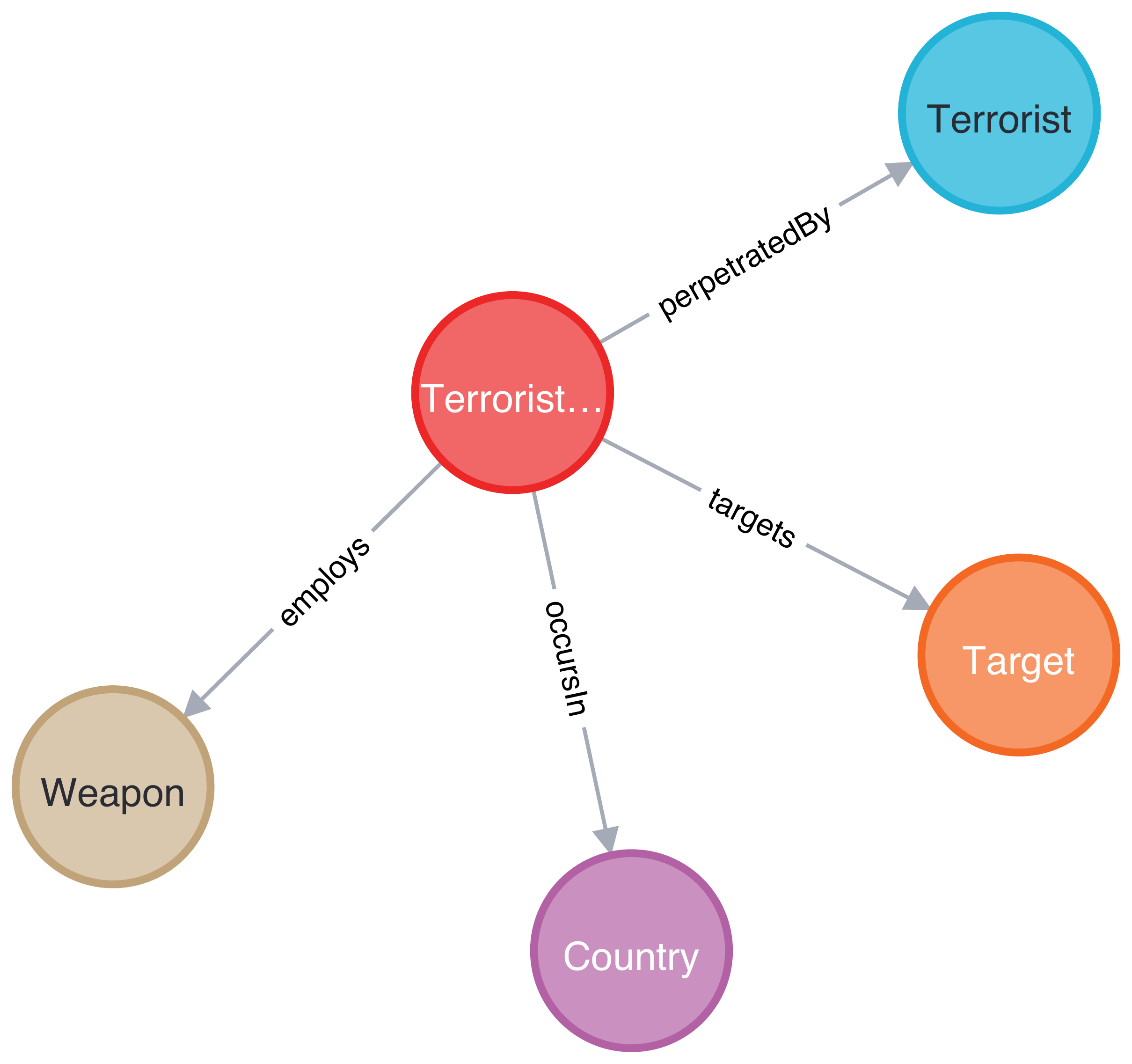}} &
\RaggedRight \tiny
\nlabel{TerroristAttack}  \nprop{number\_of\_injuries \textcolor{white}{INT}}  \nprop{number\_of\_deaths \textcolor{white}{INT}}  \nprop{name \textcolor{white}{STR}}  \nprop{locations \textcolor{white}{LIST[STR]}}  \nprop{date \textcolor{white}{DATE}} \quad  \nlabel{Weapon}  \nprop{name \textcolor{white}{STR}} \quad  \nlabel{Country}  \nprop{name \textcolor{white}{STR}} \quad  \nlabel{Terrorist}  \nprop{place\_of\_birth \textcolor{white}{STR}}  \nprop{name \textcolor{white}{STR}}  \nprop{gender \textcolor{white}{STR}}  \nprop{date\_of\_birth \textcolor{white}{DATE}}  \nprop{country\_of\_citizenship \textcolor{white}{LIST[STR]}} \quad  \nlabel{Target}  \nprop{name \textcolor{white}{STR}}
\\
\bottomrule
\end{tabular}
\end{adjustwidth}
\caption{ (Continued) Schemas of the 11 graphs in the benchmark.  }
\label{tab:all_graphs_1}
\end{table*}

\end{document}